\documentclass[journal]{IEEEtran}
\usepackage{cite}
\usepackage{color, soul}
\usepackage{dblfloatfix}
\usepackage{graphicx}
\usepackage{array}
\usepackage{caption}
\usepackage{threeparttable}
\usepackage{placeins}
\usepackage[ruled]{algorithm2e}
\usepackage{wrapfig}
\graphicspath{ {Figures/} }
\DeclareGraphicsExtensions{ .png, .pdf, .pdf}
\usepackage[cmex10]{amsmath}

\usepackage[utf8]{inputenc}
\newcolumntype{M}[1]{>{\centering\arraybackslash}m{#1}}
\title{Robust Robot-assisted Tele-grasping Through Intent-Uncertainty-Aware Planning}
\begin{document}

\author{Michael~Bowman, Songpo~Li, and Xiaoli~Zhang$^{*}$,~\IEEEmembership{Member,~IEEE}% <-this % stops a space
		\thanks{Michael Bowman is a Ph. D. Candidate in the Department of Mechanical Engineering at Colorado School of Mines, Golden, CO 80401 USA (e-mail: mibowman@mines.edu).}
		\thanks{Songpo Li is a Postdoctoral Associate in the Department of Electrical and Computer Engineering at Duke University, Durham, NC 27708 USA (e-mail: songpo.li@duke.edu).}% <-this % stops a space
		\thanks{Xiaoli Zhang is an Associate Professor in the Department of Mechanical Engineering at Colorado School of Mines, Golden, CO 80401 USA ($^{*}$corresponding author, phone: 303-384-2343; fax: 303-273-3602; email: xlzhang@mines.edu).}
		}
\maketitle

\begin{abstract}
%Promoting a robot agent’s autonomy level, which allows it to understand the human operator’s intent and provide motion assistance to achieve it, has demonstrated great advantages to the operator’s intent in teleoperation. However, research has only focused on target approaching, where we deal with the more challenging object manipulation task by advancing the shared control technique. 
In teleoperation, research has mainly focused on target approaching, where we deal with the more challenging object manipulation task by advancing the shared control technique. Appropriately manipulating an object is challenging due to the fine motion constraint requirements for a specific manipulation task. Although these motion constraints are critical for task success, they often are subtle when observing ambiguous human motion. The disembodiment problem and physical discrepancy between the human and robot hands bring additional uncertainty, further exaggerating the complications of the object manipulation task. Moreover, there is a lack of planning and modeling techniques that can effectively combine the human and robot agents' motion input while considering the ambiguity of the human intent. To overcome this challenge, we built a multi-task robot grasping model and developed an intent-uncertainty-aware grasp planner to generate robust grasp poses given the ambiguous human intent inference inputs. With these validated modeling and planning techniques, it is expected to extend teleoperated robots’ functionality and adoption in practical telemanipulation scenarios. 
\end{abstract}
\begin{IEEEkeywords}
Telerobotics and Teleoperation, Manipulation Planning, Grasping, Human Intent
\end{IEEEkeywords}
\section{Introduction}

\subsection{Need of Robot Assistance in Telemanipulation}
Teleoperating a robot allows operators to carry out tasks remotely with the robot acting as a medium for interaction. The indirect interaction by teleoperation brings advantages to the operator through increasing the motion precision and strength, access to remote work fields that might be inaccessible and hazardous. However, the indirect manipulation and physical discrepancy between human and robot hand structures often leads to difficulty and complexity in teleoperating a robot to successfully complete a task \cite{Rybarczyk2003ContributionRobot,Healey2008SpeculationSurgery}. To reduce the difficulty of controlling robots through teleoperation and the overall workload that operators face, robot agents are given more intelligence and autonomy, thus, the ability to understand operators' intended actions and assist with achieving the actions. Research has demonstrated in a target approaching scenarios robotic agents can successfully infer goal locations by observing the operator's motion trajectory which then allows the robot to provide motion assistance in approaching the target \cite{Li2015ContinuousControl,Webb2016UsingTeleoperation}. A majority of the documented research has focused on how to precisely infer target location and strategies to effectively blend human and robot trajectory inputs.

Even though the concentration of research effort has focused on the approaching process, the object manipulation task--subsequently following the approach--is an essential challenging problem that has not received enough research attention. Successfully manipulating an object requires fine motions(i.e., motion constraints for task success\cite{Song2010LearningModels}), such as grasping the object at a particular part, approaching the object in a specific angle, and applying force in in a certain manner \cite{Song2015Task-BasedInference}. The operator must accomplish these constraints using his/her own mental and physical capability to adjust the robot hand, beyond suffering from the general teleoperation disembodiment and physical discrepancy problems. Thus there is a need to enable robotic agents to proactively assist the operator in successfully telemanipulating objects for various tasks.

\subsection{Ambiguity in Robot-Assisted Telemanipulation}

One of the main barriers in intent-based control for robotic assistance is dealing with the uncertain human intent inference which could cause inappropriate robot assistance. Incorrect assistance, or task failure can result if the robot's modeling and planning process does not consider the uncertainty of human intent inference \cite{Han2016Self-ReflectiveBehaviors}. Uncertainty exists due to natural ambiguity of human motion. For instance, when a human grabs the body of a cup with a specific grasp pose in teleoperation the manipulation intent is difficult to determine from the grasp pose, since the pose itself could be for transferring the cup to another location for one operator yet be used for drinking for another operator. Additionally, indirect interaction with remote objects further increases the ambiguity of human motion and thus also increases the uncertainty in intent inference. 
\begin{figure*}[!t]
    \centering
    \includegraphics[width=\textwidth]{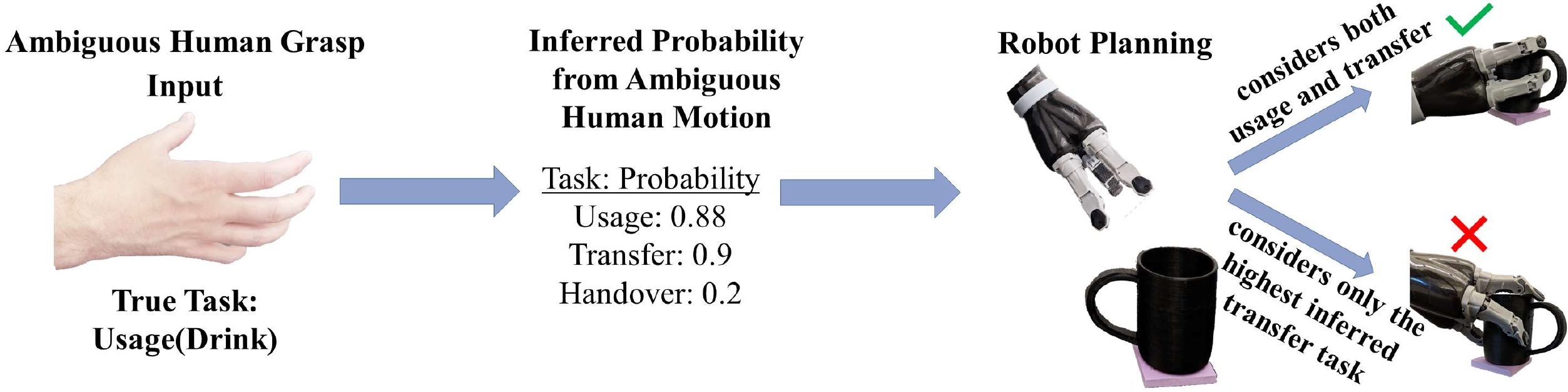}
    \caption{Intent interpretation from an ambiguous grasp lead to different solutions based upon the planning procedure. If the ambiguity in the intent is not considered it can lead to unsuccessful grasp configurations which result in the failure of the task. Considering ambiguity allows for subtle adjustments to be made to the grasp configuration to accommodate the task requirements which are close in intent and ensure the robustness of motion planning. Interpretation of ambiguous inputs and uncertainty in model outputs is crucial for the planning procedures success.}
    \label{fig:Intro_fig}
\end{figure*}

Within the ambiguous human motion subtle unique differences exist which can be used to distinguish the different tasks providing better context to humans in predicting the intended action \cite{Lin2000ModelingMotion}. These subtle differences are imperative to ensure the success of the task. For example, a perceived successful grasp for a mug is determined by task subtleties such as 1) a ``drinking" task requires sufficient room on the top and a grasping pose may dominant the handle, 2) a ``transfer to another location" task requires sufficient room on the bottom to place the object safely, and 3) a ``handover" task requires sufficient room for another person to grab which tends to create a finger dominant grasp. Although these subtleties vary between people, they still hold commonalities which are key to discern the task. Usually, these subtle differences are difficult to observe, making it harder for appropriate inference in classification models for a task, consequently provide inappropriate assistance.

\subsection{Contributions}

To solve these open issues, an intent-uncertainty-aware grasp planning method is developed for robot-assisted telemanipulation. This approach enables robot-assisted telemanipulation in which the robot infers the human operator's manipulation intent and provides grasping motion assistance to achieve the operator's intended task. In this paper, we do not focus on how to obtain human intent or manipulation intent inference, rather how to appropriately use it as an input. We also do not focus on criteria to determine what makes a successful grasping pose, or developing a new grasp model, rather our efforts focus on techniques to improve current existing grasp models. These techniques improve the models by accounting for common poses among tasks and ambiguity of human input motion and intent inference. The contributions of this work are as follows:

\begin{enumerate}

    \item Multi-Task Grasp Modeling. The modeling considers the ambiguity of human motion. Instead of building independent task models, some grasp poses share common features which result in satisfying multiple tasks. The models, therefore, contain overlap between features and poses. Additionally, non-overlapping areas are of special interest since they carry fundamentally different features from other tasks. Although these fundamental differences are often subtle, they are of critical need for the robust grasp planning model. Therefore, we developed a multi-task grasp modeling method to allow robots to understand the common features as well as subtle unique features which distinguish tasks.
        
    \item Intent-Uncertainty-Aware Grasp Planning. Given the intent inference probability input, the planner first interprets the ambiguity level of human motion, then generates the grasp based on this ambiguity level. Highly ambiguous input may need a grasp pose that comprises features from all possible tasks. Input with low ambiguity generate a grasp pose that emphasize features for the higher inferred task.
    
    \item Grasp Model Ambiguity Quantification. The approach developed is imperative to quantitatively analyze the overlap between different grasp models so designers are informed how the robot will view the multi-task grasp model. This provides insight on merging multiple grasp models and determining how similar common grasp configurations are for different tasks within the same model.
    
    \item Robustness Evaluation of Modeling and Planning. Two separate cases were analyzed to validate both the modeling and planning processes. The first case considers three separate models, which vary in degree of ambiguity overlap between tasks, to compare on the basis of task performance. The second case evaluates the task performance of two separate object-specific grasp models. 
\end{enumerate}

Our previous work \cite{Bowman2019Intent-uncertainty-awareTelemanipulationb} focused on developing a framework to deal with intent inference ambiguity(the basis for Contribution 1 and 2), yet did not go far enough in depth to dealing with other uncertainties, such as grasp model ambiguity, and object ambiguity(reformulation for a more complete framework of Contribution 1 and 2). Additionally, we provide a new understanding of ambiguity for both grasp model structure and objects(Contribution 3). A new discussion on determining the convexity of individual cases is also present(Contribution 4). There are new experimental results to show the robustness and validation of the modifications presented in the methods section(Contribution 4).

\section{Background}

\begin{figure*}[!hbt]
    \centering
    \includegraphics[width=\textwidth]{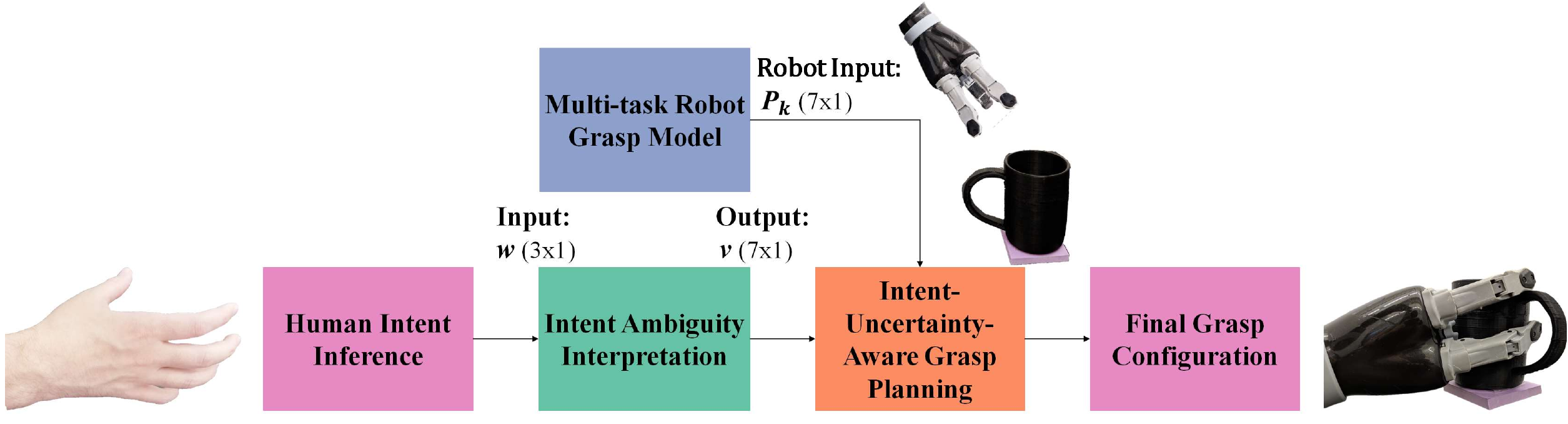}
    \caption{Demonstrates the framework used to achieve an intent-uncertainty-aware grasp planning. The initial human grasp creates a probability for each principle task, $w$. To account for ambiguity the intersection of each principle task is considered, $v$. The robot model will also account for the intersection of tasks by using common poses. Both of these produce probability vectors which are used for the intent-uncertainty-aware grasp planning. A final grasp pose is then obtained from the planning process. }
    \label{fig:block_diagram}
\end{figure*}

This section introduces formulation of shared control problems in detail by considering two modules, 1) the lack of robotic assistance and shared control for telemanipulation and 2) the challenges of framing telemanipulation as a shared control problem.
Conventional object telemanipulation is a pure master-slave strategy \cite{Leeper2012StrategiesGrasping, Corteville2007Human-inspiredMovements, Losey2018TrajectoryInteraction, Hirche2012Human-orientedTeleoperation} which relies on the operator’s tedious fine motion tuning to overcome the physical discrepancy issue between the operator hand and the robot hand, and to satisfy the subtle motion constraints for task success. This tedious master-slave strategy for complex telemanipulation brings the operator huge physical workload and mental burden, leading to task failure and user frustration. Current assistance in grasping do not use intent-based methods—nor the more needed manipulation intent—rather the current field has provided limited approaches in how to provide the assistance for firmly grasping an object without considering the context of the intended task. 

Shared control has difficulties to readily adapt to the grasping and object manipulation domain. Current assistance with (semi-)autonomous agents has focused on approaching/reaching tasks in teleoperation \cite{Khoramshahi2018AInteraction, Michelman2002SharedSystem, Kaupp2010Human-robotApproach, Mulling2015AutonomyManipulation}, however, it is not sufficient to satisfy the tele-grasping and tele-manipulation of objects. The methods to provide assistance in approaching—yet may not be work as well in grasping scenarios—include envelope motion constraints \cite{Abbott2007HapticManipulation, Webb2016UsingTeleoperation}, manually selective assistance levels \cite{Feygin2002HapticSkill, Li2003RecognitionFixtures}, and shared control policies such as linear blending \cite{Aarno2005AdaptiveTasks, Dragan2013AControl}. Linear blending strategies may not entirely work as the motion constraints from the manual operator’s perspective and the fully autonomous perspective may differ. Alternatively, intent-based shared control has shown promise and improvement in approaching tasks by placing the burden of task completion on the autonomous system. These approaches in providing assistance work for approaching because there is no need to be concerned with the physical discrepancy, largely ignore fine motion constraints, and no additional modeling requirements. Therefore, these methods used in approaching tasks may be ill-suited for grasping tasks with regards to task success. 

Task success is considered a higher-level goal than grasp success. A grasp can be considered successful based on criteria for grasp stability and contact pressure \cite{Huebner2012BADGr-AGRasping, Cutkosky1989OnTasks, Quispe2016GraspingPlanning}, however, just because a grasp is successful does not mean it is appropriate. For instance, if one were to be at a tea party where it is socially acceptable to drink in a particular manner, others may be confused if you do not conform to their style. There may be many ways you could successfully grab a teacup to drink; however, it would result in a task failure if others do not perceive it as appropriate. Thus, one can think of task success applying context for a grasp success. Within the same vein of task success for grasping, additional robot models (dependent on hand structures) are needed to successfully achieve the task. Especially, in shared control techniques for teleoperation, this task-level “style” or “planning” is determined by the human operator. The physical discrepancy in end-effector design requires different poses to be generated from grasp models.  For instance, the poses generated using the same intent inference for a two-finger robot and five-finger robot may be totally different from a human pose to achieve the same intent. 

Additionally, in a teleoperation scenario the robot specific task constraints may differ from the human task constraints to accomplish the task. Likewise, the robot specific task models may conflict with human input commands as shown in Fig. \ref{fig:Intro_fig}. Although the poses could be drastically different from the human input motion, yet subtle common constraints exist which can be exploited for task success. For example, when transferring a cup to another location the human may want to grasp it from the body, while the robot grabs it from the top, both of which fundamentally leave enough room on the bottom to place the cup. This is especially true of object manipulation as common grasp configurations may be able to solve multiple tasks because in practical scenarios where uncertainty and ambiguity exist, and situations cannot be entirely disambiguated. The robot agent should instead understand and apply common task constraints which are satisfiable between the two agents.

\section{Methods}

\begin{table*}[!bhp]
    \centering
    \caption{SUBTLE FEATURE COMPARISON OF COMMON POSES WITHIN A MULTI-TASK MODEL FOR A CUP}
   \begin{tabular}{|M{0.75cm}|M{1.9cm}|M{1.9cm}|M{1.9cm}|M{1.9cm}|M{1.9cm}|M{1.9cm}|M{1.9cm}|M{1.9cm}|}
   \hline
   Task&Usage&Transfer&Handover&Usage and Transfer&Usage and Handover&Transfer and Handover&All\\
   \hline
    Top View&\includegraphics[width=1.6256cm]{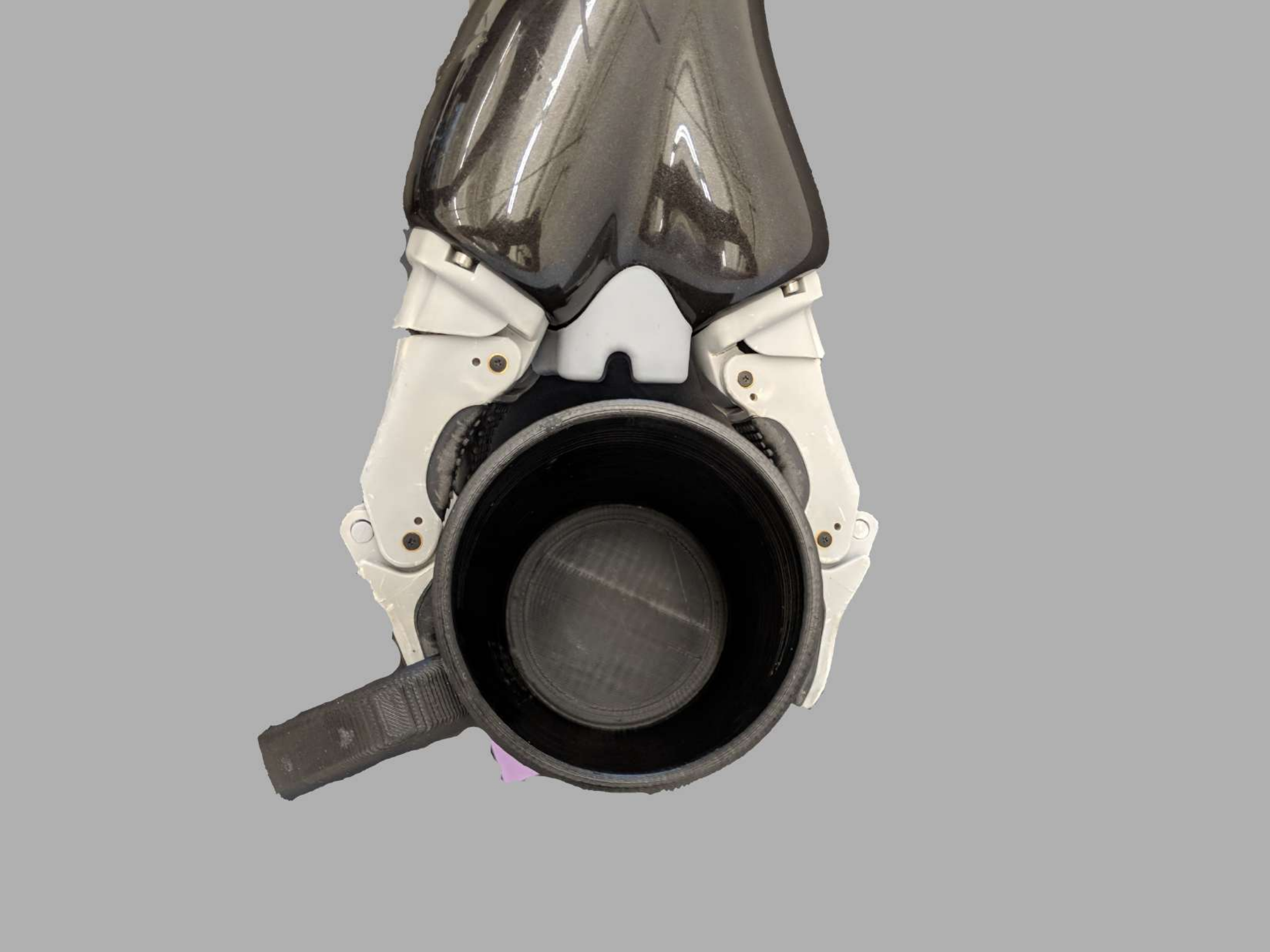}&\includegraphics[width=1.6256cm]{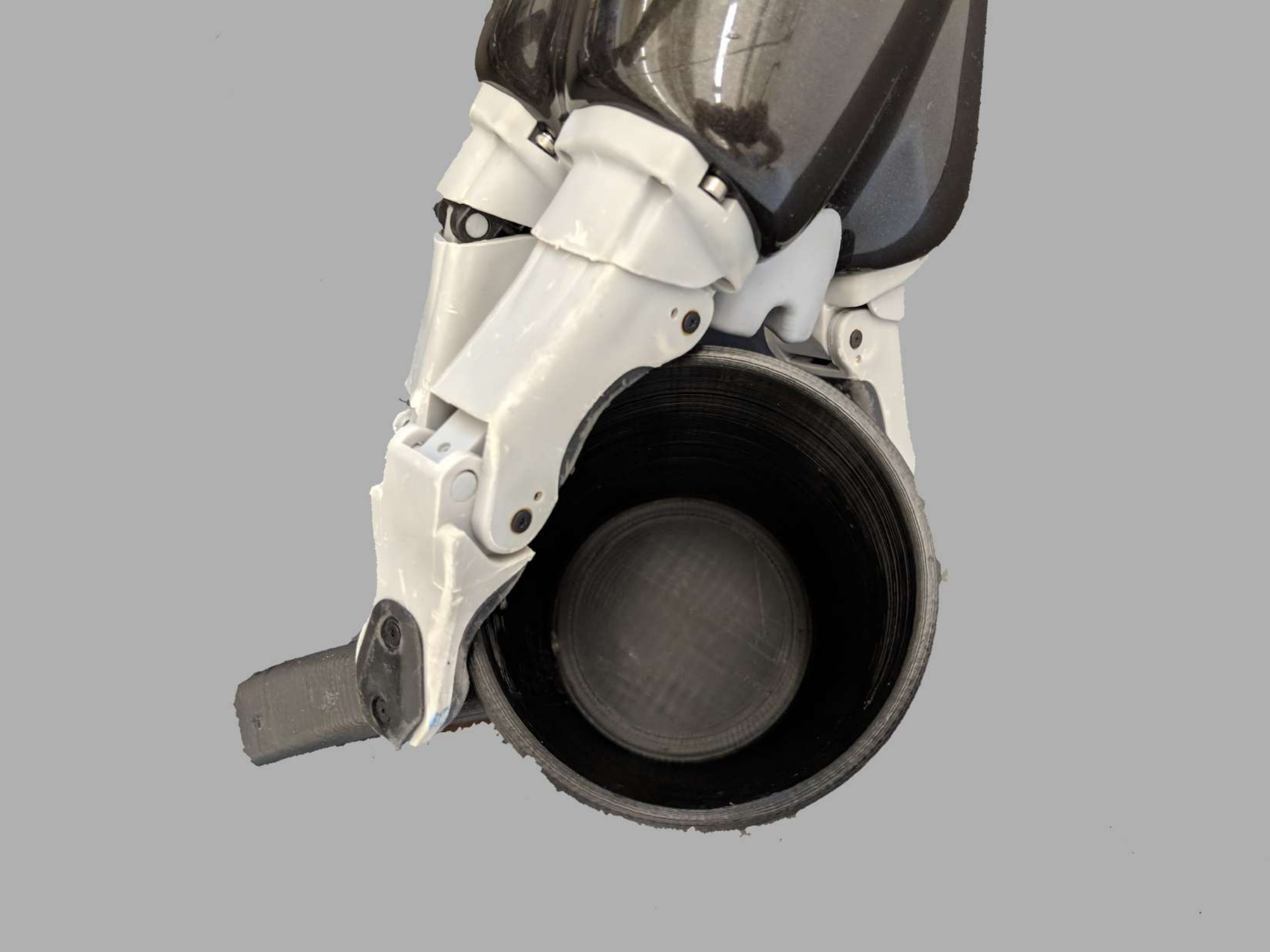}&\includegraphics[width=1.6256cm]{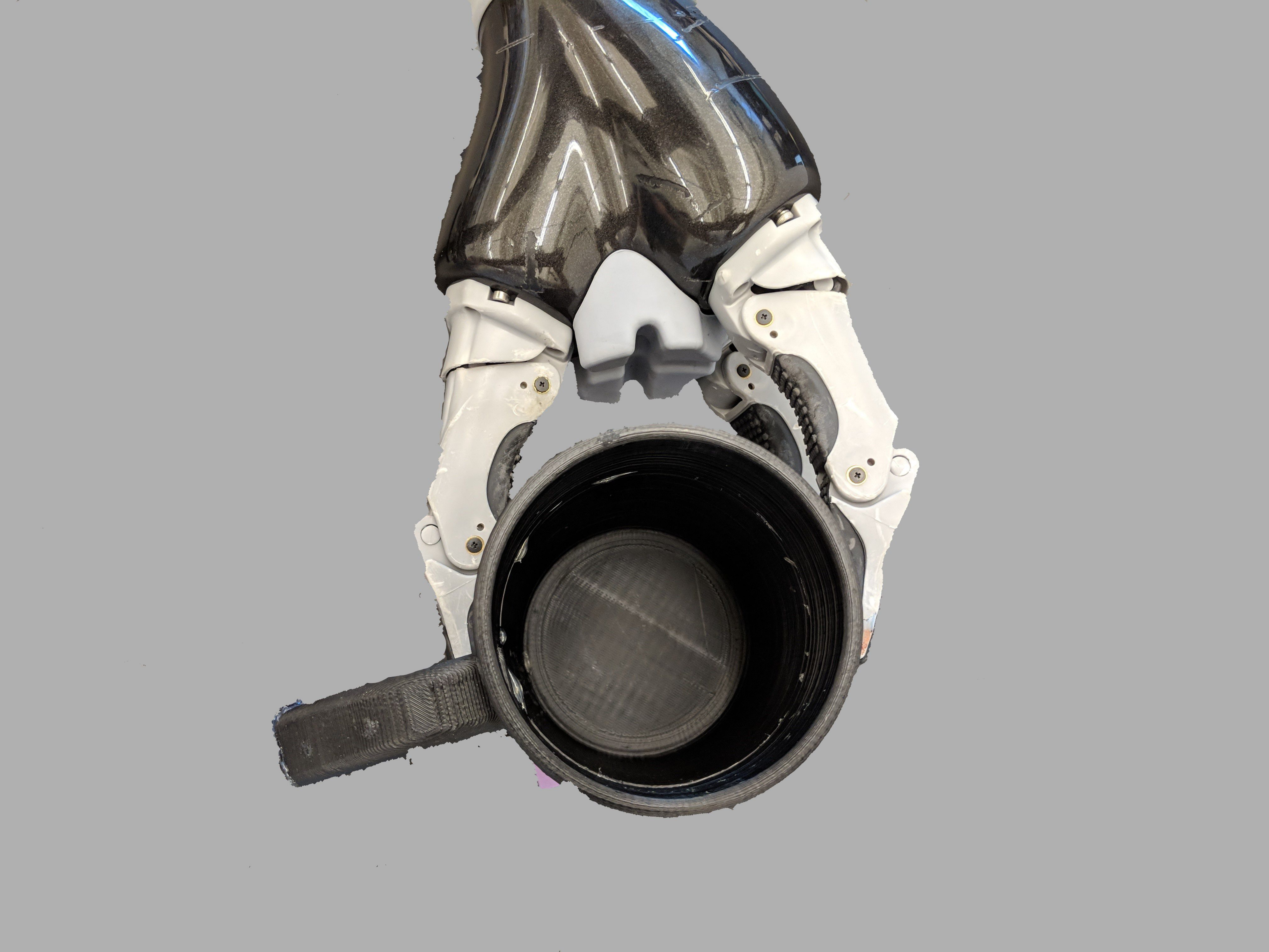}&\includegraphics[width=1.6256cm]{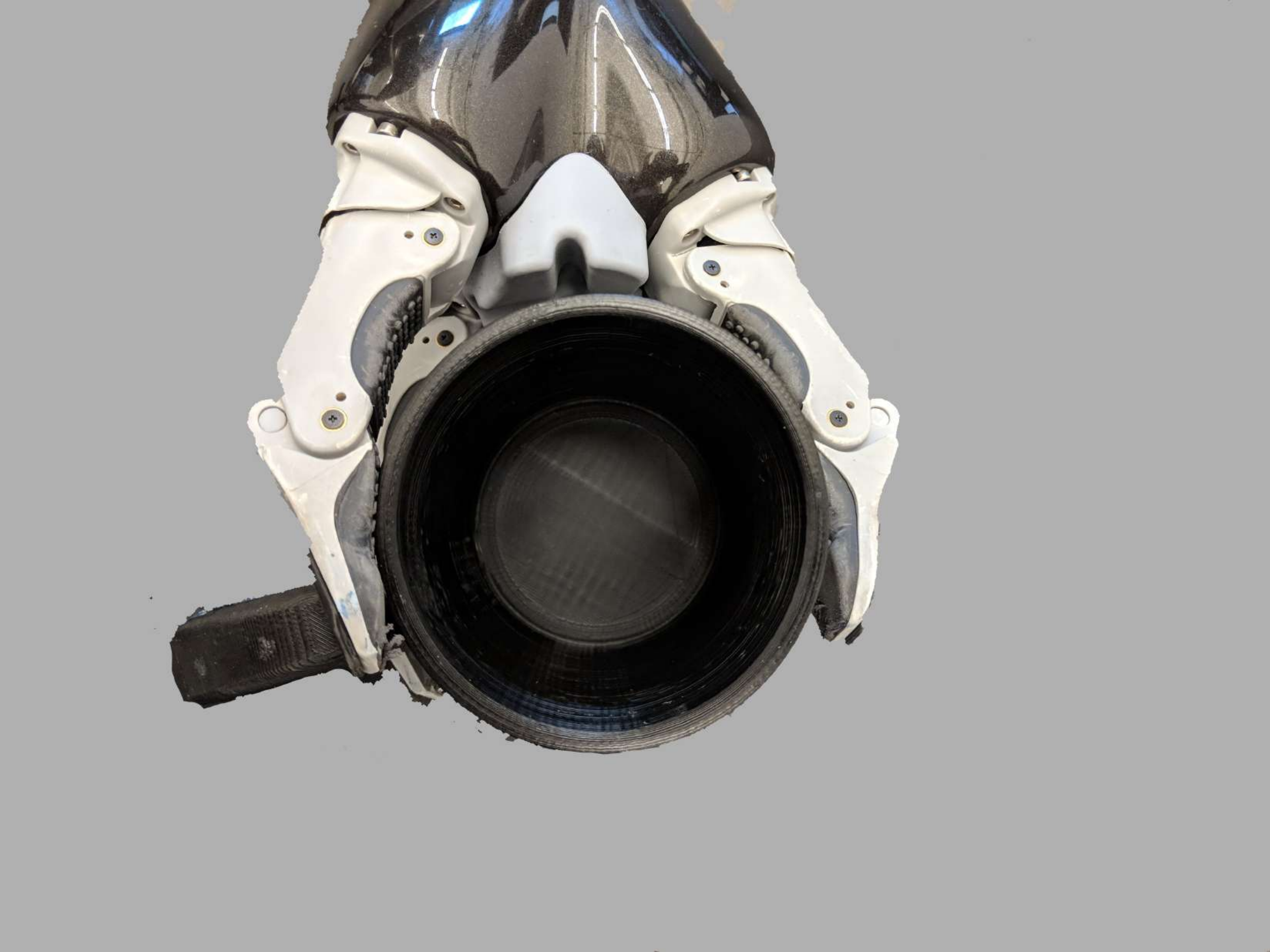}&\includegraphics[width=1.6256cm]{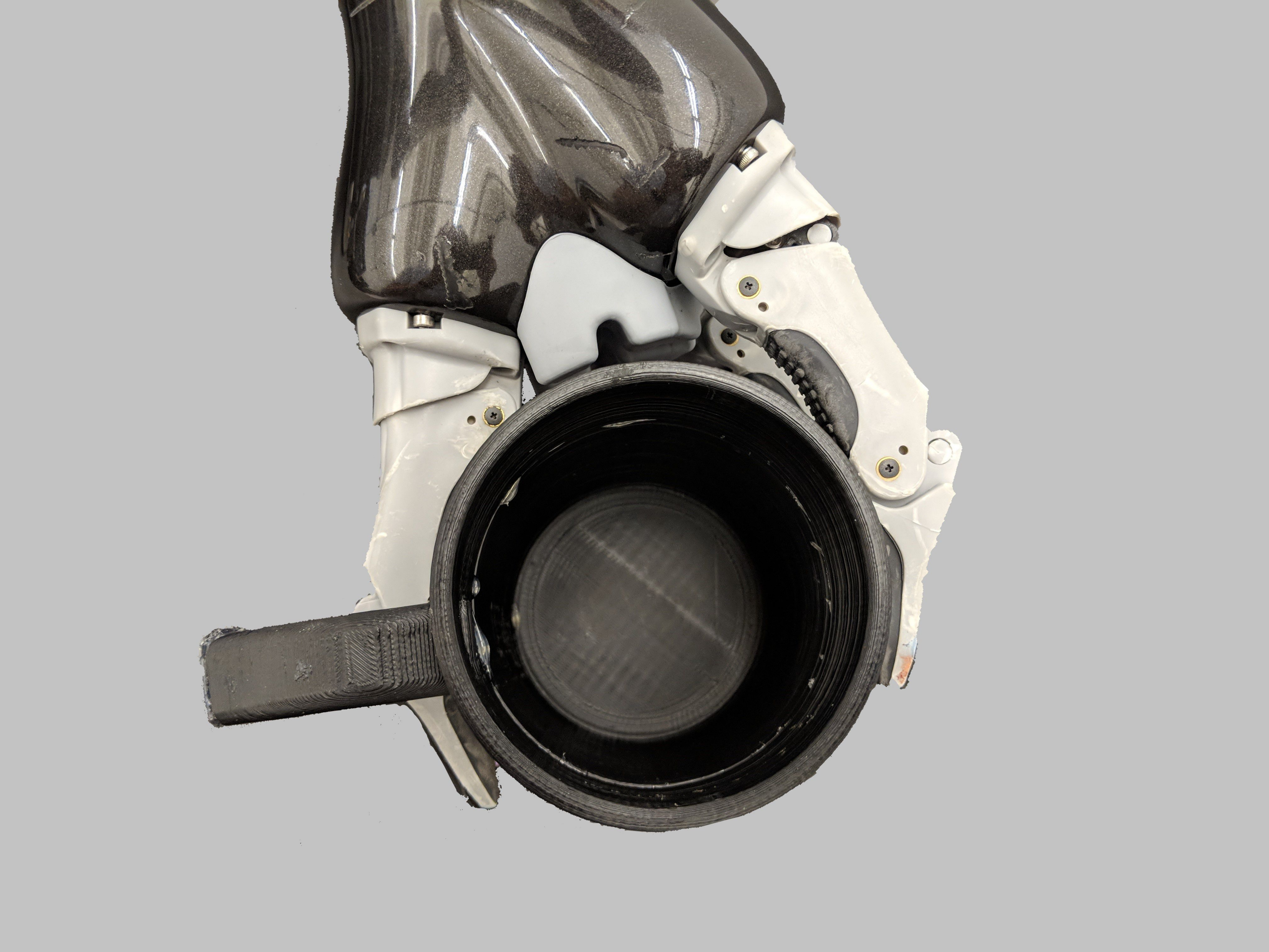}&\includegraphics[width=1.6256cm]{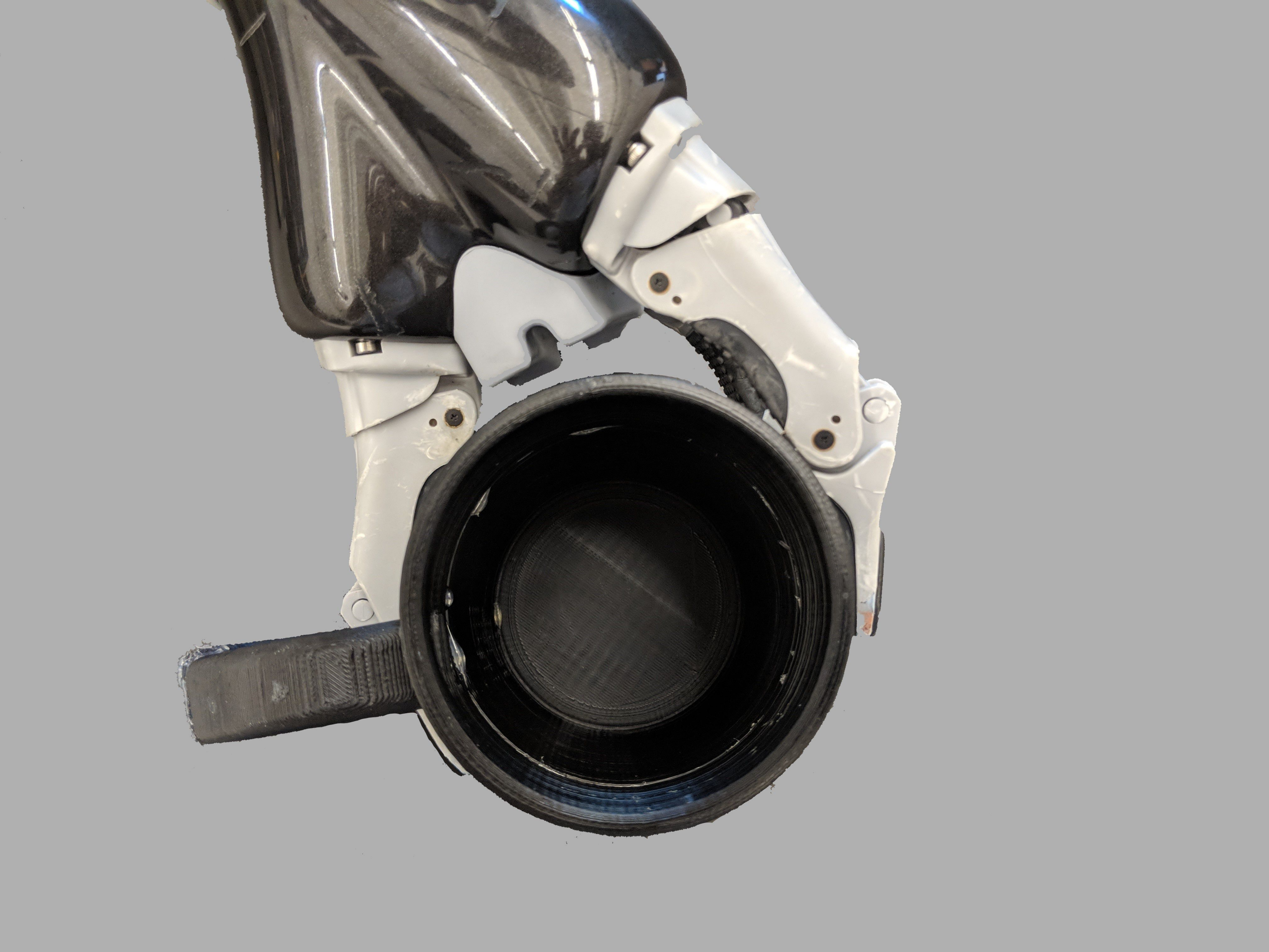}&\includegraphics[width=1.6256cm]{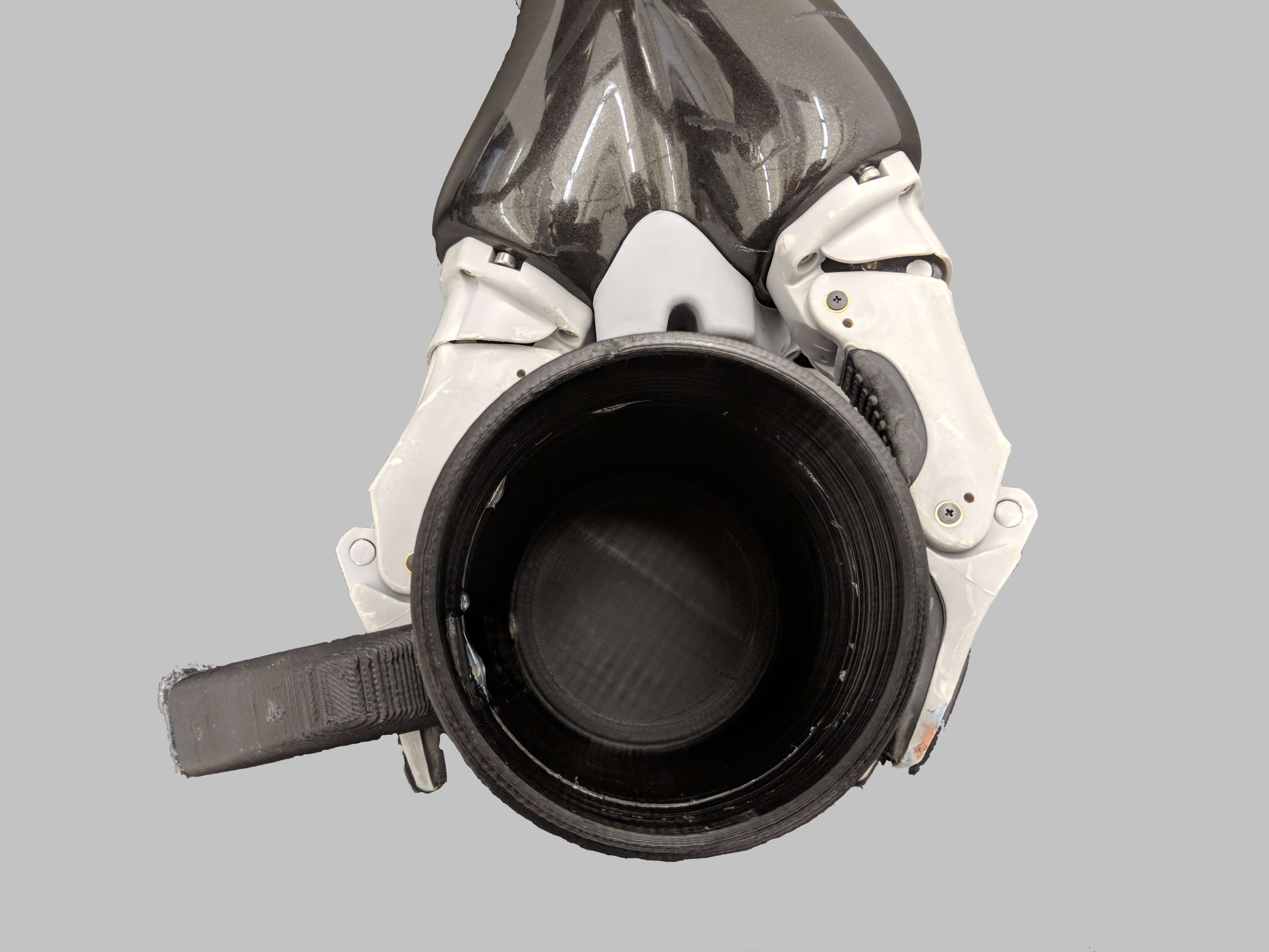}\\
    Left View&\includegraphics[width=1.6256cm]{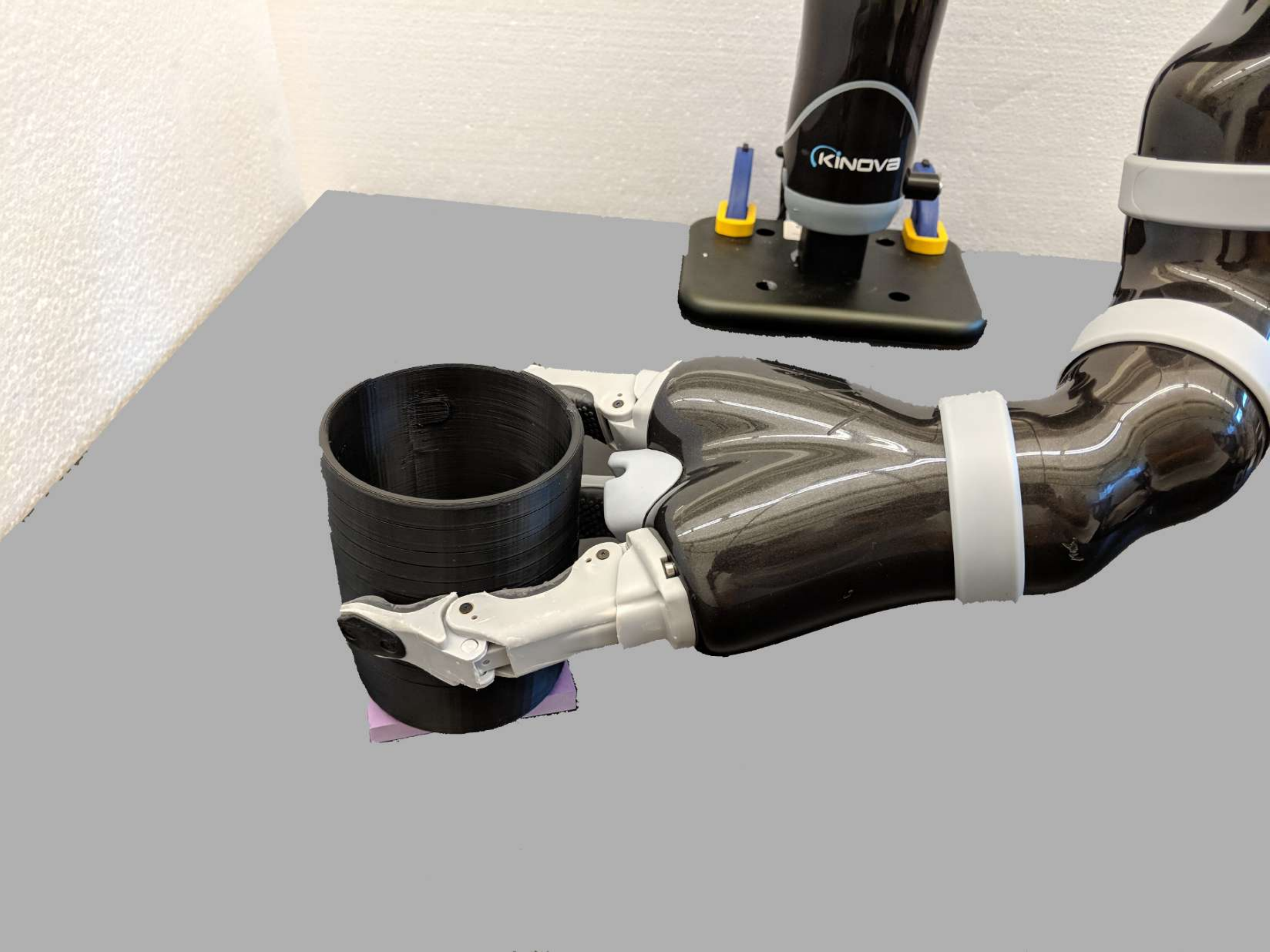}&\includegraphics[width=1.6256cm]{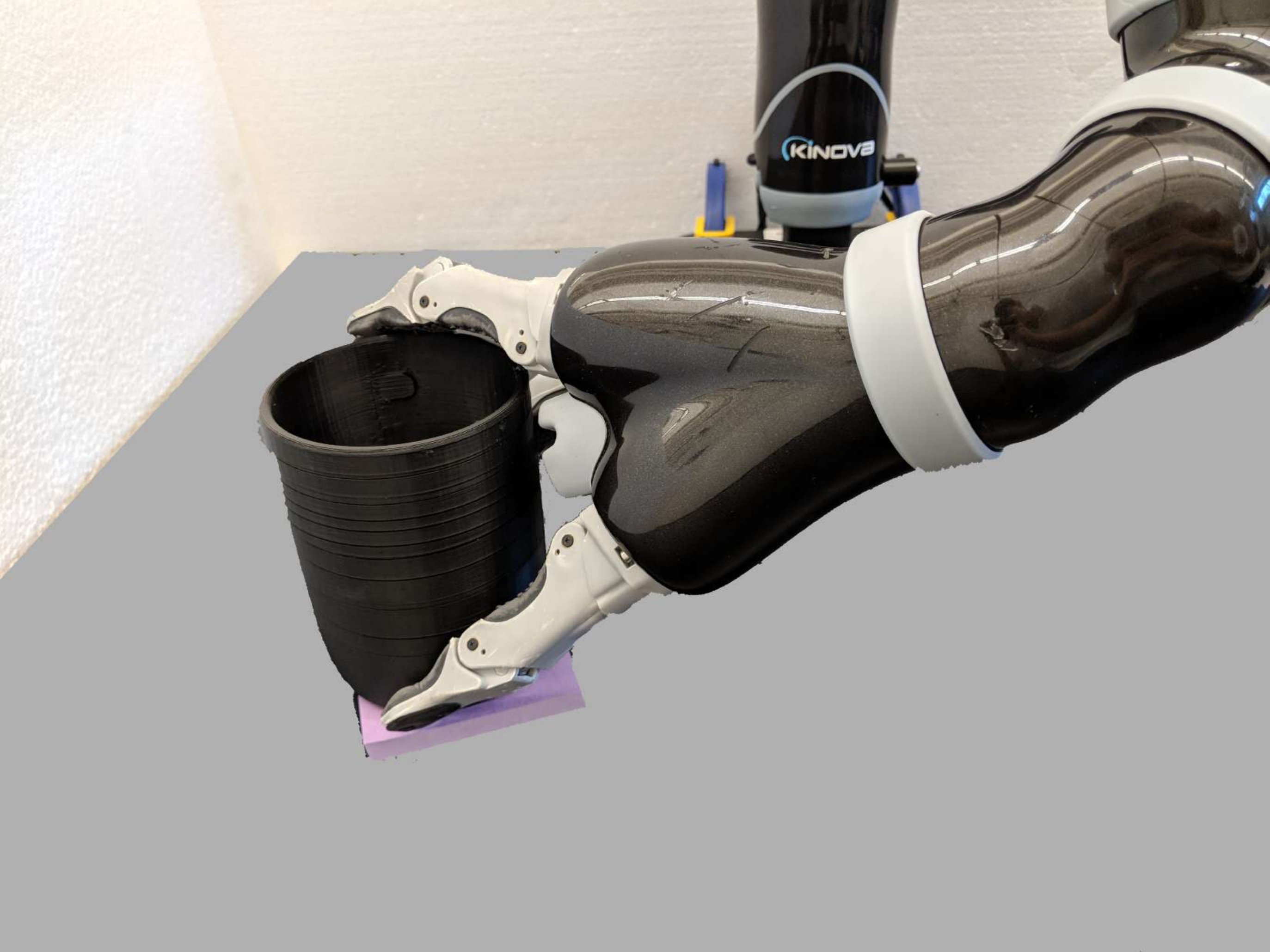}&\includegraphics[width=1.6256cm]{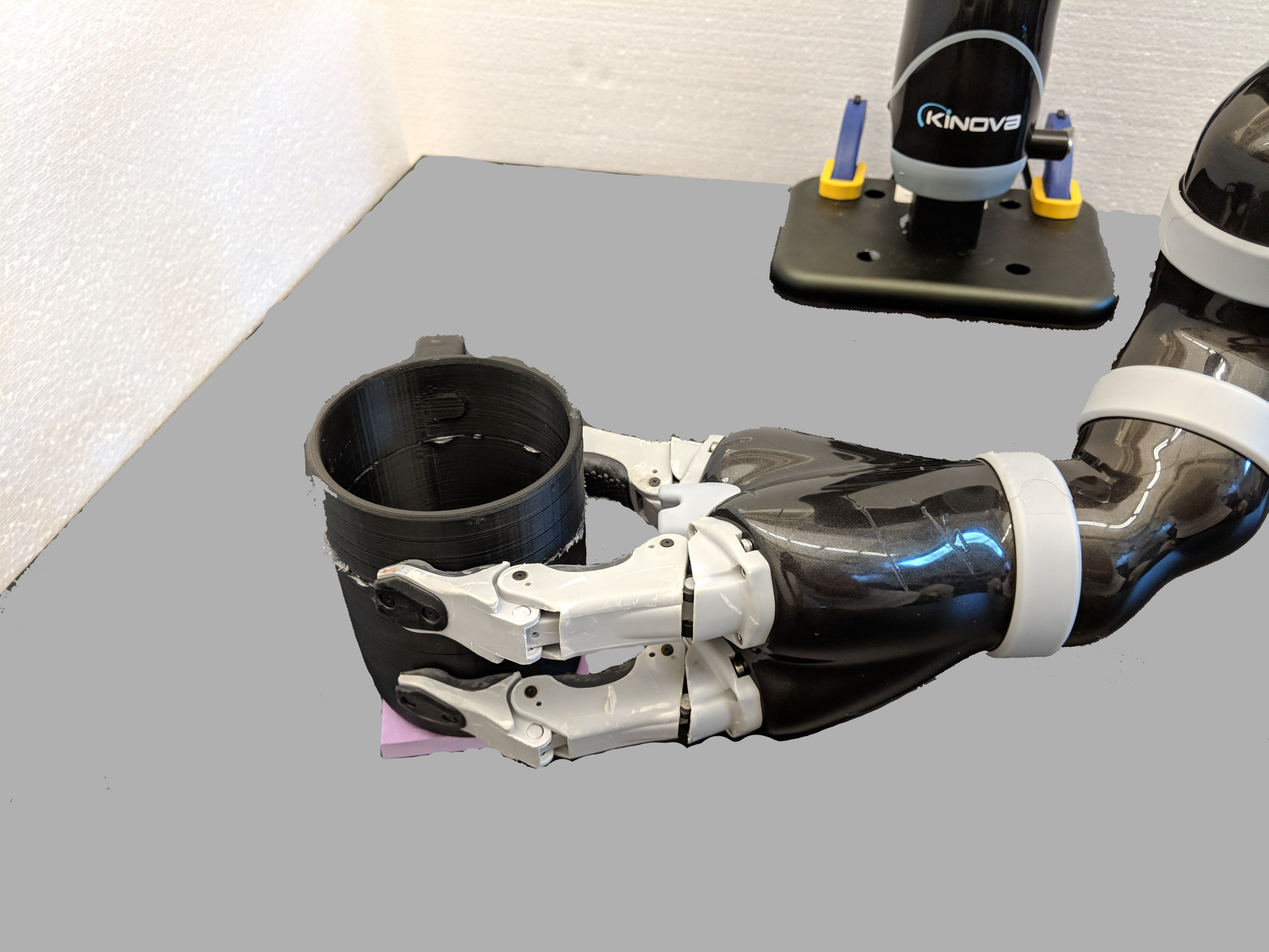}&\includegraphics[width=1.6256cm]{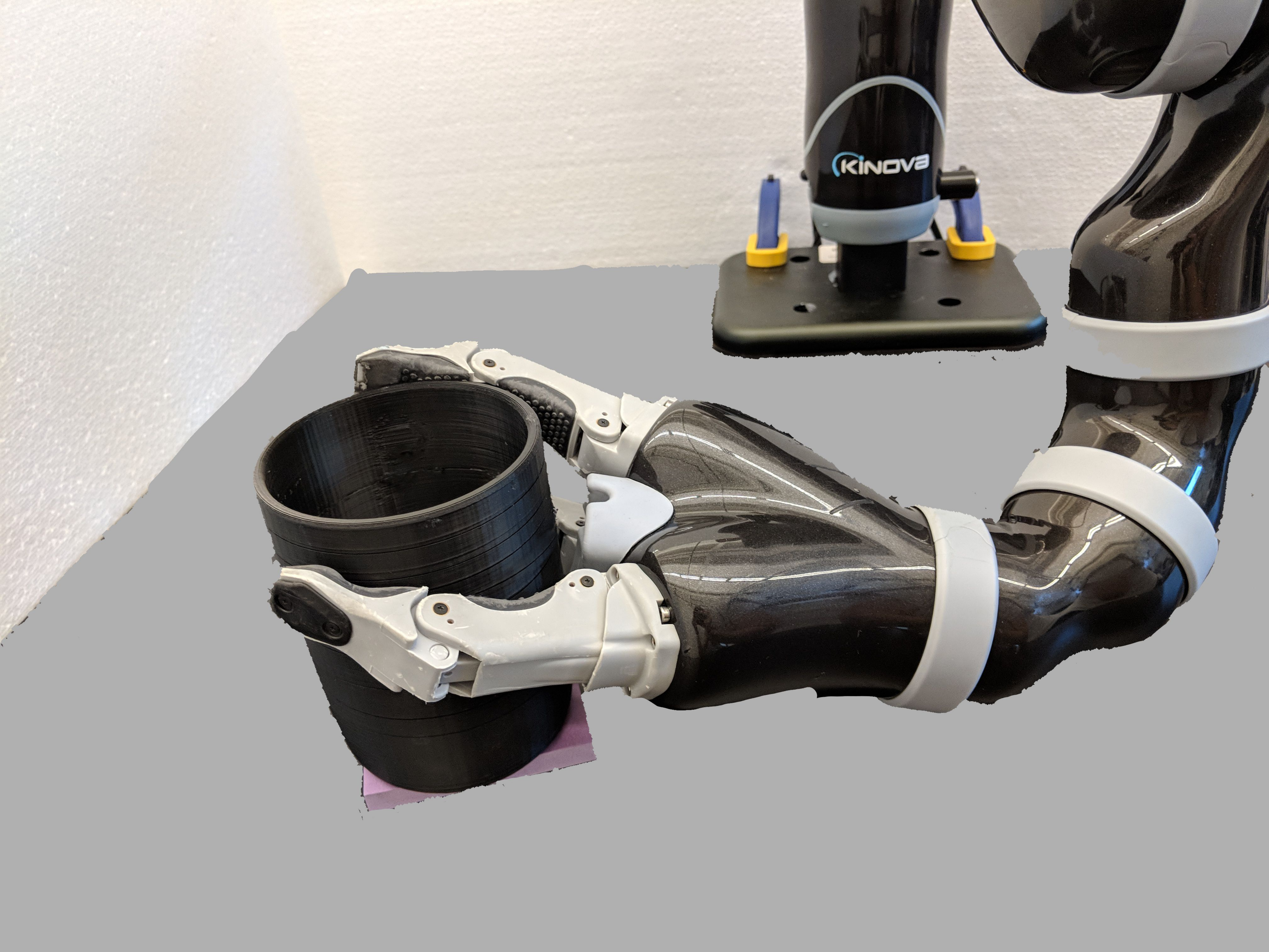}&\includegraphics[width=1.6256cm]{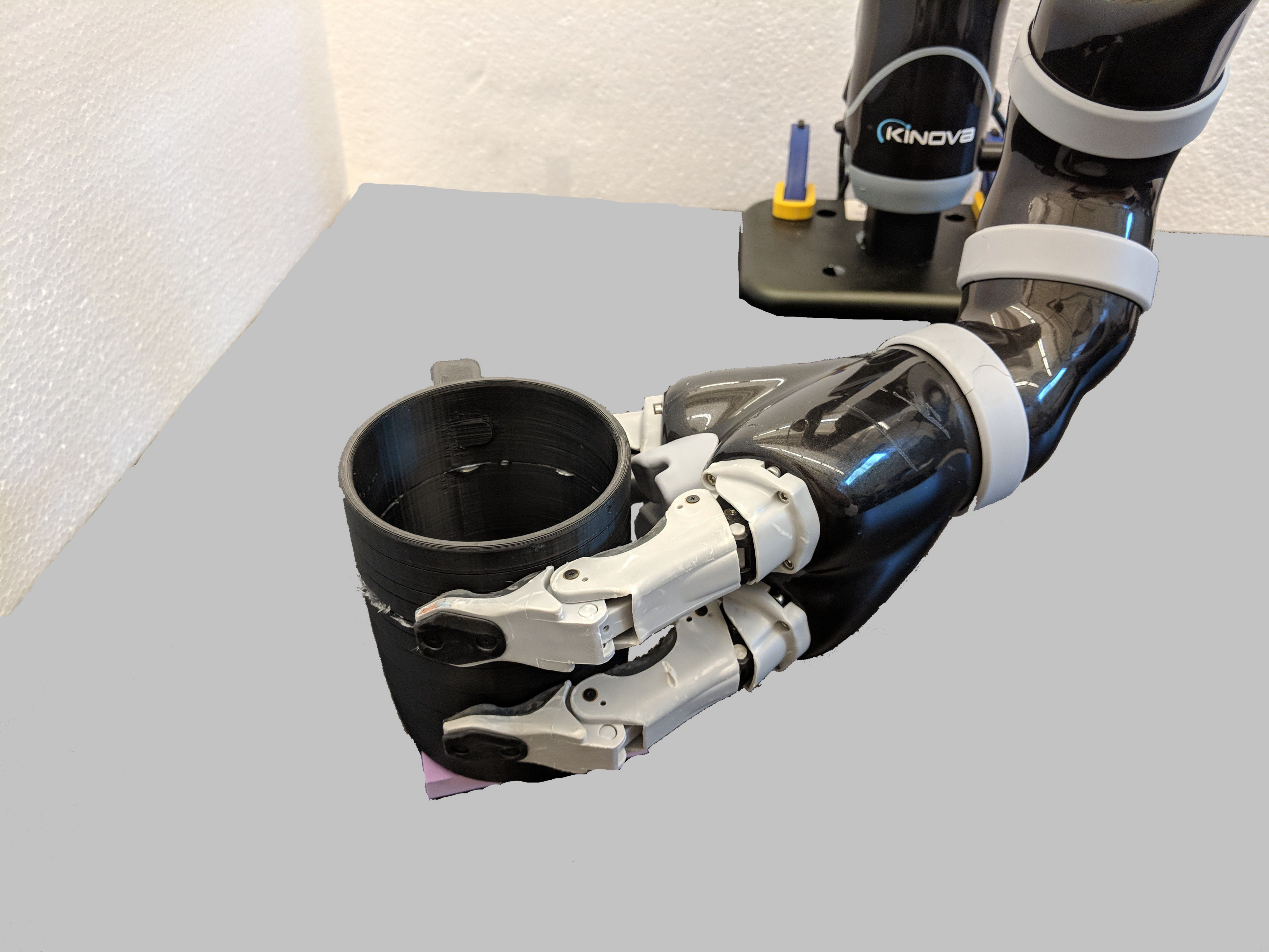}&\includegraphics[width=1.6256cm]{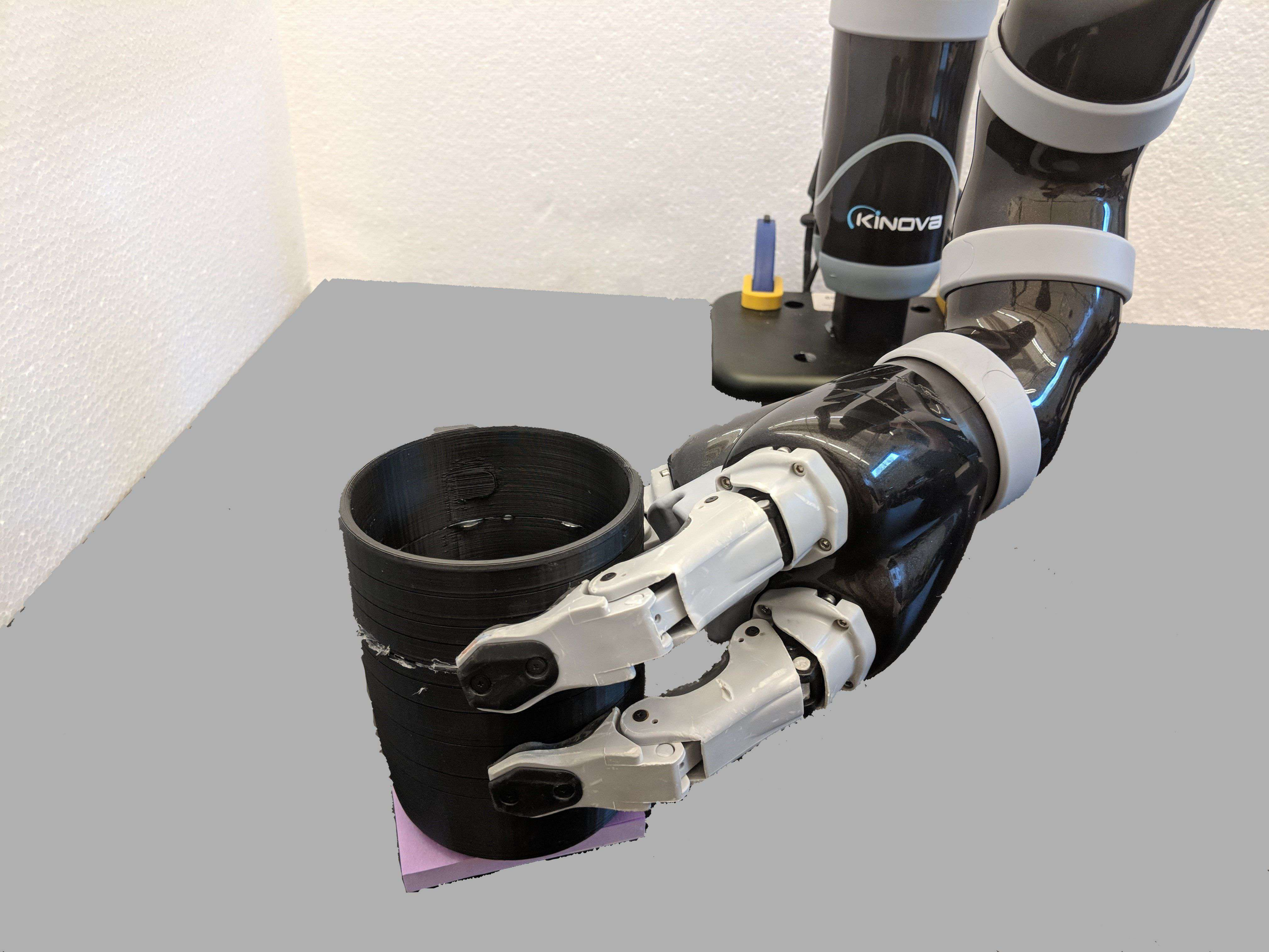}&\includegraphics[width=1.6256cm]{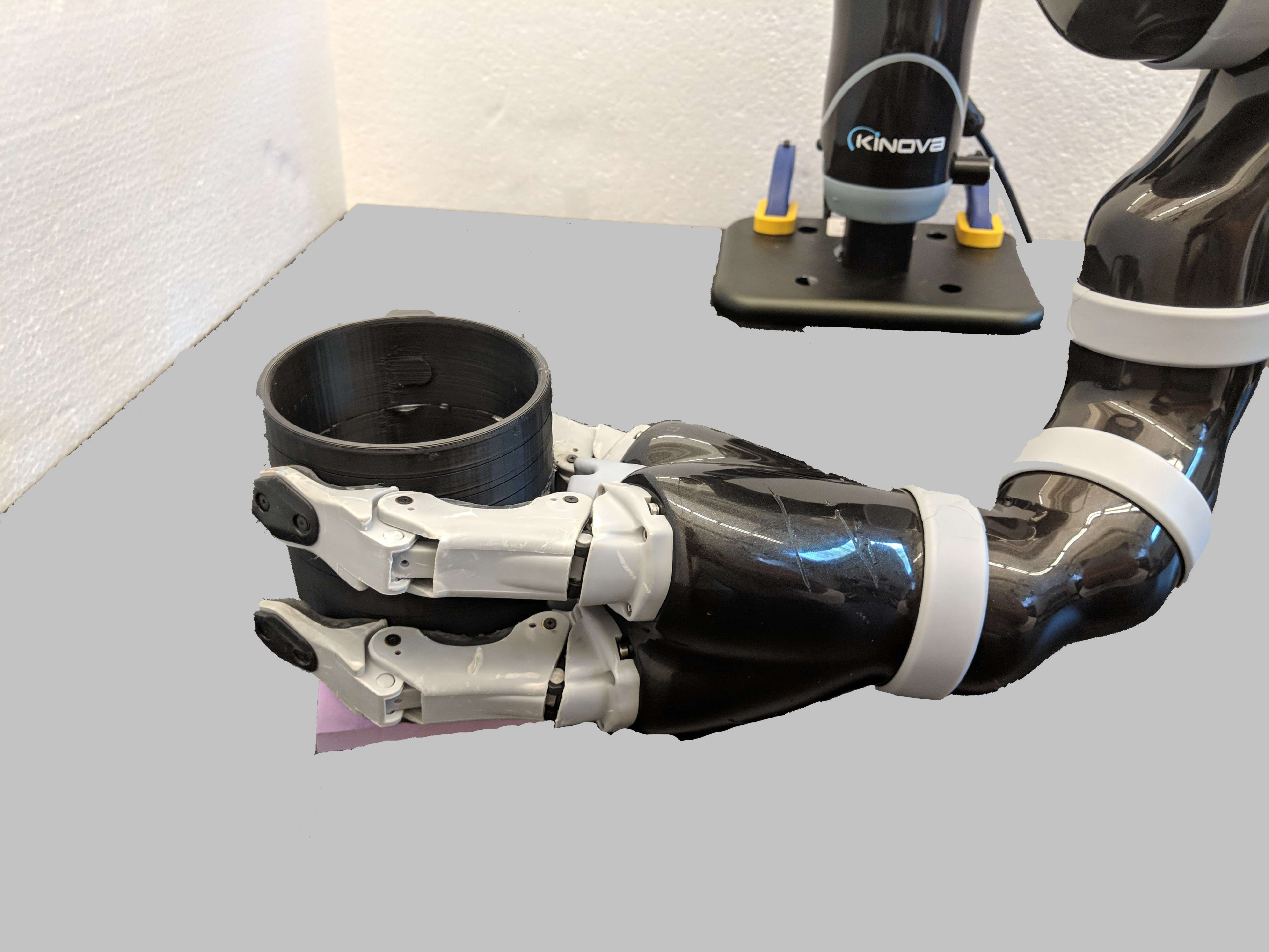}\\
    \hline
    Subtle Detail&More fingers to dominate handle space. Lower palm to leave sufficient room on top to drink.&Prefers to be perpendicular to table and higher on cup to ensure space on bottom to place cup.&Lower palm, leaves room for another user to grasp handle ($180^\circ$ rotation of wrist).&Fingers dominate the handle. Palm position allows room to drink and place cup.&Few fingers in handle space but occupies more of it. Lower palm to leave room to drink.&Few fingers in the handle space for user. Higher palm to ensure space on bottom.&Few fingers in the handle space for user. Enough space to drink and place.\\
    \hline
   \end{tabular}
    \label{tab:Cup_subtle_table}
\end{table*}{}

\subsection{Framework Overview}
\begin{figure}[!tbhp]
    \centering
     \includegraphics[width=\columnwidth]{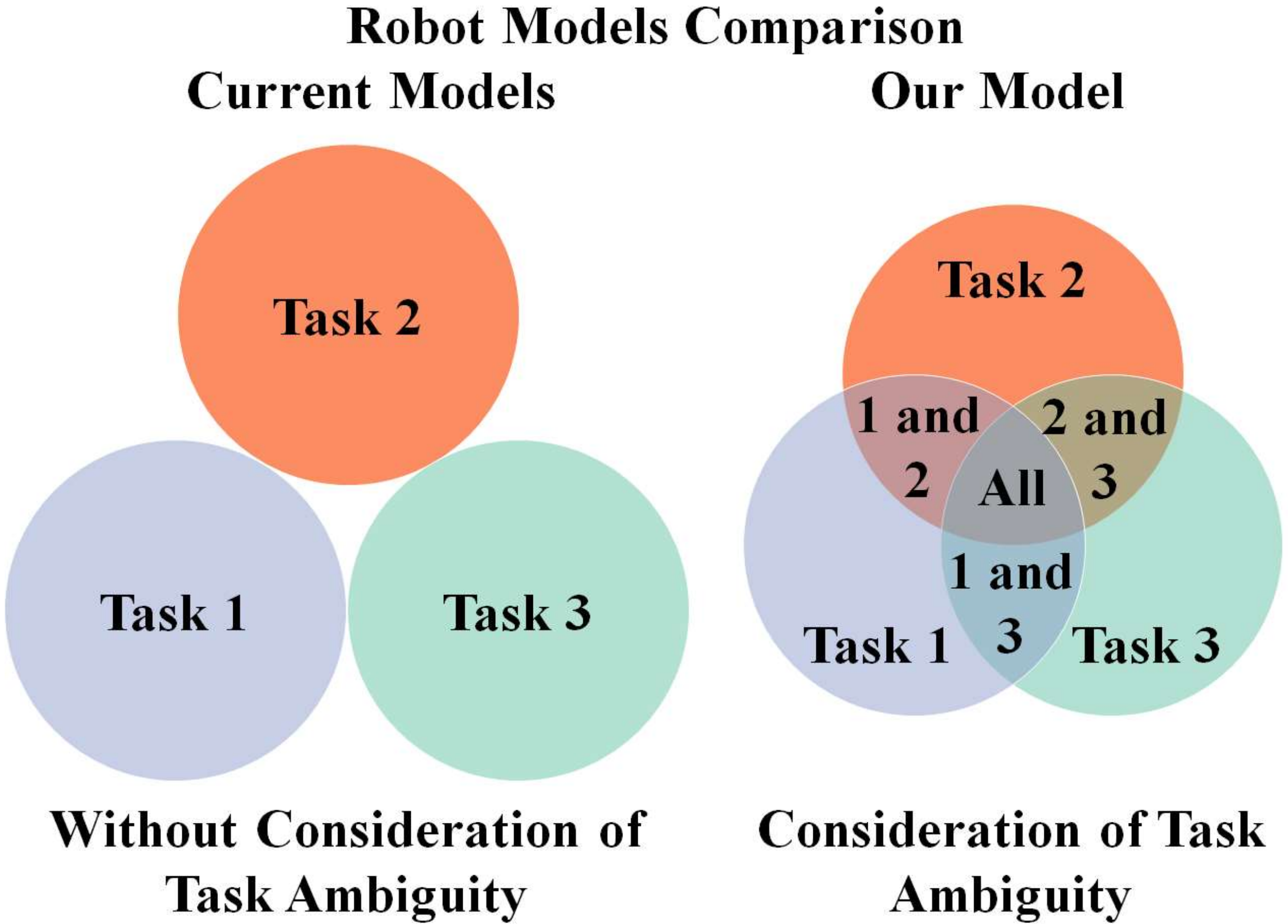}
     \captionof{figure}{Robot models need to consider task ambiguity. Current models(left) do not consider ambiguity or where multiple tasks may be satisfied simultaneously. An alternative model(right) considers tasks being concurrently fulfilled by common poses.} \label{fig:Circle_Plot}
\end{figure}
The overall framework flow can be seen in Fig. \ref{fig:block_diagram}. The focus of this section is on three main components, the multi-task robot model, the human intent ambiguity interpreter, and the intent-uncertainty-aware grasp planner. The multi-task robot model considers the overlapping nature of different tasks due to common motion features shared by their grasp poses and allows the robot to understand the fundamental different features of these tasks. Grasping objects is ambiguous without a clear discrete difference between satisfying specific tasks because of the inherent nature of the manipulation problem where this varies between people. Thus by interpreting the human intent inference, a descriptor for the ambiguity level among the tasks can be provided which the robot will attempt to satisfy. The planner attempts to find a robot pose by pulling characteristics from different tasks depending upon the ambiguity levels among tasks.

\begin{table}[!htbp]
    \centering
     \captionof{table}{INCLUSIVE LABELING FOR PRINCIPLE TASKS WITH CONSIDERATION OF COMMON POSES}\label{tab:inclusive_label_tab}
     \includegraphics[width=\columnwidth]{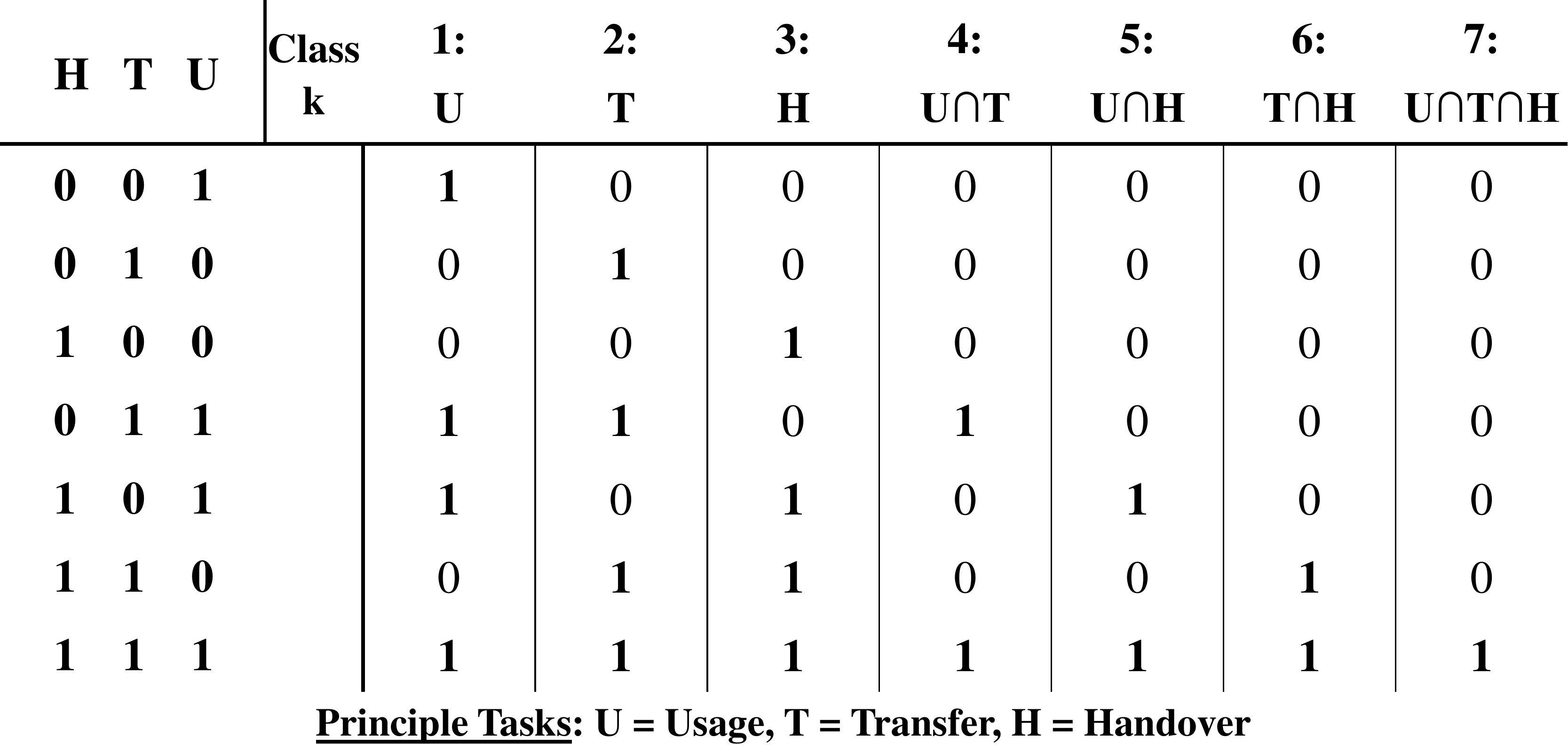}
\end{table}{}

\subsection{Multi-task Robot Modeling}
Unlike more traditional models, we propose a model structure which uses common poses in conjunction with probability distributions, naturally this leads to a Bayesian network design. Traditional Bayesian structures \cite{Song2011MultivariateGrasping, Song2013PredictingInteractions}, as shown in Fig. \ref{fig:Circle_Plot}, do not consider ambiguity \cite{Zhou2007SolvingClustering} of task distinctions, where our method on the right shows the consideration for overlapping tasks. Current robot models create task models separately from one another--to create clear, independent distinctions between tasks--and lack consideration of common poses. This causes difficulties to find continuous features between poses. Common poses have a unique ability to satisfy multiple tasks, however, they may be sub-optimal to satisfying a single principle task and less transparent to independent models. However, traditional Bayesian structures can obtain probability rather easily \cite{Cheng2013ComparingClassifiers}. To ensure the probability for our model is correct it is imperative the training data is labeled appropriately and reflects the right side of Fig. \ref{fig:Circle_Plot}.

\subsubsection{Inclusive Labeling}

Common poses add a degree of ambiguity compared to those of independent model strategies. Although, our approach takes into consideration the ambiguity by creating a multi-class Bayesian structure \cite{Tzima2015InducingSystems,Zhang2010Multi-labelDependency}. Our model considers this by giving an inclusive aspect where poses which satisfy more specific situations should be included in the more general cases. For instance, in Fig. \ref{fig:Circle_Plot}, all tasks within the Task 1 circle should be considered within the Task 1 class. The example model created was for grasping a cup--as shown in Table \ref{tab:Cup_subtle_table} in conjunction with Fig. \ref{fig:Circle_Plot}--for principle tasks which include: Task 1, usage or drinking, Task 2, transfer to another location, and Task 3, handover to a person.

Compared with traditional grasp modeling techniques, the classes and labeling of our modeling method will not be the three principle tasks, exclusively, rather they will contain seven classes in which the robot can satisfy. To label these classes correctly, the assumption made is a robot pose which satisfies more principle tasks than necessary is admissible to a more general class because it satisfies the necessary principle task. By creating an inclusive class structure, the model can begin to understand the subtle and unique differences among tasks. For instance, when looking at Table \ref{tab:Cup_subtle_table}, we observe a group labeled for the class ``Usage and Transfer". Poses falling under this label satisfy not only the class ``Usage and Transfer", but also the individual classes ``Usage" and ``Transfer", thus should be considered in the training sets for ``Usage" and ``Transfer" tasks. This duplication, or information sharing, of training sets allows the more inclusive class (such as Task 1, Task 2, and Task 3) to identify the distinctions between unique and common features. Table \ref{tab:inclusive_label_tab} shows how to appropriately label and consider the training set for the model created by the right of Fig. \ref{fig:Circle_Plot} in conjunction with Table \ref{tab:Cup_subtle_table}. The letters U, T, and H stands for the principle tasks, ``Usage", ``Transfer", and ``Handover" respectively. The columns represent each class the robot model knows, while the rows are the strict subset of events which makes up each class. A pose which is used for ``Usage and Transfer only"(row 4), can satisfy both ``Usage"(column 1) and ``Transfer"(column 2) alone as well as ``Usage and Transfer"(column 4). All poses which could be used to satisfy a ``Usage" task (column 1) include, ``Usage only"(row 1), ``Usage and Transfer only"(row 4), ``Usage and Handover only"(row 5), and ``All Tasks"(row 7).

When observing Table \ref{tab:Cup_subtle_table}, we notice between the classes the pose features are subtle, yet they are critical. By labeling the training poses in this manner, we train the model to identify subtle differences and common poses for the principle tasks as well as unique specific cases within the principle tasks. Subtle features can be difficult to observe and determine, however, they provide better context to a human user. Table \ref{tab:Cup_subtle_table} includes the subtle features used for our model including high level concepts such as palm contact, clearance from the top of the cup, and sufficient room for a user to grasp. These subtleties are further discusses in the table.

\subsubsection{Multi-task Modeling}

The goal of the Bayesian model is to obtain the posterior probability, or the probability of each class given a continuous pose \cite{John2013EstimatingClassifiers, Heckerman2008ANetworks}. Each pose contains continuous features, $x$, which can include position, orientation, force, and object features. Our model included palm center, and palm orientation as well as finger force. The model is a Naive Bayes classifier (NB), but uses the features to create a multivariate normal distribution as shown in \ref{eq_one}. The key parameters from the model for each classification zone, $k$, include: the mean value of each feature, $\mu_k$, the covariance matrix, $\Sigma_k$, and the prior probability, $P(k)$. These parameters are learned from the training set by using the Expectation Maximization Algorithm. In (\ref{eq_one}), it shows the conditional probability of $x$ given $k$, where $d$ is the length of $x$.
\begin{equation}\label{eq_one}
P(\boldsymbol{x}|k)= \frac{1}{\sqrt{ \det|\boldsymbol{\Sigma_k}| (2\pi)^d}}e^{-\frac{1}{2}(\boldsymbol{x}-\mu_k)^{T} \boldsymbol{\Sigma_k}^{-1} (\boldsymbol{x}-\mu_k)}
\end{equation}
With this equation, Bayesian models obtain the posterior probability of the class $k$ given $x$ with (\ref{eq_two}), where the class with the highest probability is the label for $x$.
\begin{equation}\label{eq_two}
    P(k|\boldsymbol{x}) = \frac{P(\boldsymbol{x}|k)P(k)}{\sum_{k}^{K}P(\boldsymbol{x}|k)P(k)}
\end{equation}
Each probability of a class is then put into what is referred to as the \textit{robot probability vector, $P_k$}. The model we created from Fig. \ref{fig:Circle_Plot} and Tables \ref{tab:Cup_subtle_table} and \ref{tab:inclusive_label_tab}, means there are seven classes and the probability for one pose to satisfy all seven classes must sum to one.

\subsection{Intent Ambiguity Interpretation}

To handle the ambiguity, the modeling of human manipulation intent follows a concept of multi-label classification, where two, three, or $m$ principle task classification model outputs could be satisfied at the same time, shown in Fig. \ref{fig:Intro_fig}. Each classification model produces a probability and is put into a vector of $m$x1 size where this is referred as the \textit{classification input vector, $w$}. Since each model output is independently obtained, we can identify each joint probability case. This will generate a vector of size $2^m$ where $m$ is the number of principle tasks. This vector of possible events with their subsequent probability will be referred to as the \textit{human probability vector, $u$}, shown in (\ref{eq_three}), where $\Psi (m)$ is the power set from 1 to $m$ task classification models, and Y is each subset of $\Psi (m)$.
\begin{multline}\label{eq_three}
P(\bigcap_{i\in Y}w_i \bigcap_{j \notin Y, j \in m}\neg w_j)=\\
\prod_{i\in Y} P(w_i)\prod_{j\notin Y,j\in m}1-P(w_j), Y\subset \Psi(m)
\end{multline}
For example, consider when m=3 as shown in Fig. \ref{fig:Intro_fig}. The classification probability vector(0.88,0.9,0.2) produces a vector where each combination of the principle tasks is considered true and false. So when event $U\cap \neg T \cap \neg H$ occurs, the probability is $0.88(1-0.9)(1-0.2)=0.07$, while when the event $U\cap T \cap \neg H$ occurs, the probability is $0.88(0.9)(1-0.2)=0.63$.

Since the robot model is unaware of the event where inaction should be taken($\neg U\cap \neg T\cap \neg H$), it is important to eliminate this action and normalize the rest of the probability vector where at least one principle task $m$ is satisfied. This is to ensure the robot classes, $k$, is equal to $2^m-1$. This new vector is known as the \textit{target probability vector, $v$}, and is shown in (\ref{eq_four}). The target probability vector now has a scenario associated with each class of the robot model previously discussed in Tables \ref{tab:Cup_subtle_table} and \ref{tab:inclusive_label_tab}.
\begin{equation}\label{eq_four}
    v_k = \frac{u_k}{\sum_{j}^{2^m-1} |u_j|} ~~~~\forall_k < 2^m
\end{equation}
Alternatively, a designer could force the robot to ask for clarification if the inaction task is sufficiently high. The human probability vector has a distinct advantage over using the classification intent by allowing more descriptive behavior of the type of features the robot model should use. For instance, Fig. \ref{fig:Intro_fig} shows if the classification intents for grasping a cup for Usage and Transfer are similar and high while the Handover task is low, then the decision process should carry features which demonstrate and use features from both high intents. Additionally, the human probability vector inherently accounts for uncertainty in human intent by allowing for the system to consider multiple classes at once as well as varying degree the multiple classes may influence the outcome of chosen features. 

\subsection{Intent-Uncertainty-Aware Grasp Planning}

The planning process involves taking the ambiguity levels developed in the intent interpretation section and using them as a basis for the determining characteristics needed. Whether the ambiguity levels are high or low will determine which set of features should predominantly be used from the robot model classes developed earlier. The planning process will start from a known pose, which is closest to the ambiguity levels, then continue to refine the features of the pose until a more reflective pose is created. The refining step is iterative and will attempt to use both unique and subtle characteristics from all classes as it needs until convergence.

The planning process includes using a Bayesian structure to best fit the target probability vector. Traditional Bayesian models attempt to maximize the probability of one region or class over all others, since these models are based on independent model strategies, rather than minimize the difference between the current and target probability vectors. By matching the probability vectors, the subtle and unique features of the robot model should reflect the best combination of intent to make it predictable to the user. The first step is to identify the grasp class out of the defined zones and select an initial grasp pose from the training set of this class, since this is currently the best approximation the robot model can make. This best approximation is done by taking the highest class of the human probability vector and choosing the best pose which is reflected in the class of the robot model. Determining the best pose within the class can be done by using the objective function as shown in (\ref{eq_five}). The objective function is a least square regression which will attempt to match the robot probability vector to the target probability vector.
\begin{equation}\label{eq_five}
    min ~~~ \frac{1}{2} \sum_k (v_k - P_k)^2
\end{equation}
By mimicking the probability, the unique and subtle features of the robot model will reflect those most similarly to the human intent. This is due to the robot model classes containing both the unique and subtle features which the planner takes characteristics. The next step is to refine this initial approximation and determine how the features need to be adjusted. To determine the amount of adjustments the planner will take, from a unique feature to a more common feature or vice versa, a gradient descent method was applied as shown in (\ref{eq_six}).
\begin{equation}\label{eq_six}
    \frac{\partial P_k}{\partial x}(v_k - P_k)~~~~ \forall_k
\end{equation}
This can be solved using finite-difference methods \cite{Waltz2006AnSteps} or analytically using matrix calculus \cite{BrandtPetersenMichaelSyskindPedersen2012TheCookbook}. This approach will use the ambiguity levels as a guide to determine which subtle or unique features are more critical, or influential in ensuring a successful grasp pose. Afterwards, constraint equations will be applied--which are robot and case dependent based upon features selected thus will be left to the reader's discretion--and the objective function will be evaluated once again. Algorithm \ref{Algorithm_1} shows the process where all together these pieces will provide the best features to match the human intent.
\begin{algorithm}
\SetAlgoLined
\caption{Intent-Uncertainty-Aware Grasp Planning}\label{Algorithm_1}
\BlankLine
\KwResult{ \textbf{ $x$, $P_k$ }}
\BlankLine
        \underline{\textbf{Sets and Indices:}}{
        
        i: Feature of grasp configuration 1...I\
        
        k: Task of each grasp configuration 1...K\
        }
        
        \underline{\textbf{Parameters:}}{
        
        $U_i$: Upper bound on feature i\
        
        $L_i$: Lower bound on feature i\
        
        $v_k$: Target probability for task k\
        }
        
        \underline{\textbf{Variables:}}{
        
        $x_i$: robot input for feature i\
        
        $P_k$: Probability function of satisfying task k given robot features x
        
        }
        
        \underline{\textbf{Objective:}}{
        
        ~   $min ~ ~\frac{1}{2}~\sum_k (v_k-P_k)^2$\
        
        ~       ~   s.t. ~ ~$L_i \leq x_i \leq U_i ~ ~\forall_i$\
        
        ~   ~       ~~~~~~~$1=norm(x_i)~ ~ \forall_i$ needed for palm direction\
        }
\end{algorithm}
Despite being able to find suitable candidates which can satisfy human intent, this approach cannot guarantee convexity. This is due to three key factors. First, convexity is dependent upon the Bayesian Network structure. For instance, a more interconnected Bayesian structure would mean not all inputs are independent of one another adding to the nonlinearity of the model. Along these lines, a multivariate normal distribution was used, however, more appropriate distributions may be required for each feature within the structure. The distributions may not compliment one another by being from the same distribution family like the multivariate normal distribution. Second, by adding the inclusive labeling, more information is added to these specific distributions to encompass more parts to the unique tasks. The addition of the inclusive labeling adds more potential candidates which may cause the overall distribution function to span a greater range and potentially increase the nonlinearity. Lastly, since this approach is built on data driven modeling, thoughtful data collection is necessary to ensure sensible solutions and bounds. 
\subsection{Determining the Degree of Model Ambiguity}

An important aspect to address with the approach is to evaluate the degree of overlap between multiple tasks. We propose to use Kullback-Leibler(KL) divergence \cite{Hershey2007AppoximatingModels} method for grasp model ambiguity. The goal of this divergence is to be able to determine how different two populations of data are from one another. For this to work, we assume one population is the true population, and the other is the inference to see how well we can use one to predict the other. However, two populations of data, may diverge from one another differently. For instance, if a large population contains a subset population, being able to predict the large population from the subset may be easier than determining what the subset does given the larger population. Therefore we establish there is a true population, and an inference population which will be denoted by P and Q, respectively. It should be noted the KL divergence is bounded [0,$\infty$]. Where if there is no divergence (where P is used to infer itself) the value is 0. In the context of modeling the principle task ambiguity, this means if the divergence is 0, there is maximum ambiguity between the two principle tasks. Therefore the more the divergence that exists the more disambiguated the tasks appear. 

\begin{multline}\label{kl_eq}
KL(\boldsymbol{P}||\boldsymbol{Q})= \frac{1}{2}(Tr(\Sigma_Q^{-1}\Sigma_P)+(\mu_Q-\mu_P)^T\Sigma_Q^{-1}(\mu_Q-\mu_P) \\
-d+\ln{\frac{|\Sigma_Q|}{|\Sigma_P|}})
\end{multline}

For example, if we were to attempt to infer the Transfer task from the Handover task, we would assign the P as the Transfer task, and Q as the Handover task, where the $\mu$'s and $\Sigma$'s for each task would be assigned. This can be done for each combination of two tasks. By placing the combinations in a matrix we obtain a nonsymmetric matrix.

\begin{equation} \label{kl_mat}
    \begin{matrix}
    0 & KL(U||T) & KL(U||H)\\
    KL(T||U) & 0 & KL(T||H)\\
    KL(H||U) & KL(H||T) & 0\\
    \end{matrix}
\end{equation}

The matrix shows the differing degree of overlap, where eigen values can be obtained to determine the most divergent task model. Alternatively, the nonsymmetric divergence matrix can be used to determine the upper bounds on of the statistical distance using Pinsker's inequality.

\begin{equation}\label{pinsker_eq}
    \delta = \sqrt{\frac{1}{2}KL(P||Q)}
\end{equation}
The statistical distance is another metric to determine how similar the populations are to one another where KL divergence provides an upper bound on each aspect of the task model. Statistical difference can also be put in matrix for easier comparison and determining which tasks are the most divergent. 

However, a symmetric matrix is another possible way to quantify the divergence. This is a fundamentally different way to analyze the overlap as the goal would be to look at the total amount of divergence between the two tasks.
\begin{equation}\label{kl_sym_mat}
\resizebox{\linewidth}{!}{$
    \begin{matrix}
    0 & KL(U||T)+KL(T||U) & KL(U||H)+KL(H||U)\\
    KL(U||T)+KL(T||U) & 0 & KL(T||H)+KL(H||T)\\
    KL(U||H)+KL(H||U) & KL(T||H)+KL(H||T) & 0\\
    \end{matrix}$}
\end{equation}

The nonsymmetric and symmetric matrices analyze different aspects of the overlap, where the nonsymmetric matrix is more component oriented, while the symmetric matrix focuses more on the overall model overlap. The necessity of using one or the other is algorithm and case dependent, however, both can provide insight on the behavior of the model. Despite these metrics being used to determine the overlap, it is susceptible to errors due to these methods being data dependent. The data collected and used for the grasp modeling is crucial and primarily dominates the behavior of ambiguity.

\section{Results}

\subsection{Experimental Setup}
The experimental setup included using a Kinova Mico arm as well as two objects (a coffee mug and a flashlight) for interaction. The robot model was gathered for three tasks: using the object(i.e drinking from the mug or shining the light on a workspace), handing the object over to another person, and transferring the object to another location. Each model is built separately and only considers features for the final grasp configuration and did not include trajectory or temporal features, however, the force in which the fingers applied were considered. Both models were created by using expert analysis of rules to determine which task could be satisfied. The rules developed express extreme cases where there are certain cases of no overlap between principle tasks. The robot was then manually moved to obtain the training data. To solve the formulation, two separate solving techniques, sequential quadratic programming(SQP) \cite{Nocedal2006NumericalOptimization} and Interior-Point \cite{Waltz2006AnSteps}, were used, however, other solvers may be used to achieve quicker convergence. These two solvers were used to compare the overall solution with one another to ensure the same local minimum was achieved. A comparison is not discussed as both solutions converge on the same solution at nearly the same time. For a more in depth comparison between these specific strategies see \cite{Boggs2000SequentialOptimization}.

Three separate objective experiments are used to verify and validate the techniques. The first is to analyze \textit{inference ambiguity}, by using a single object model where we compare simulated classification intent. The three cases of intent we compare are when one task is dominant, two tasks are codominant and where all three are equal, respectively. The second experiment analyzes \textit{model ambiguity}, by comparing different degrees of ambiguity or overlap by looking at different principle models. The different models consider a differing number of classes, where there are cases where some unique and no unique grasp configurations exist. The third experiment is to analyze \textit{object ambiguity}, aiming to compare different objects using the same intent inference as an input. Additionally, there are two subjective tests to determine the effectiveness of the techniques with human subjects. The first determines the subtlety and compare the rules of the grasping model to the final generated grasp pose, while the other compares human predictability to the classified grasp.

\subsection{Objective Results}

\begin{figure*}[!b]
    \centering
    \includegraphics[width=\textwidth]{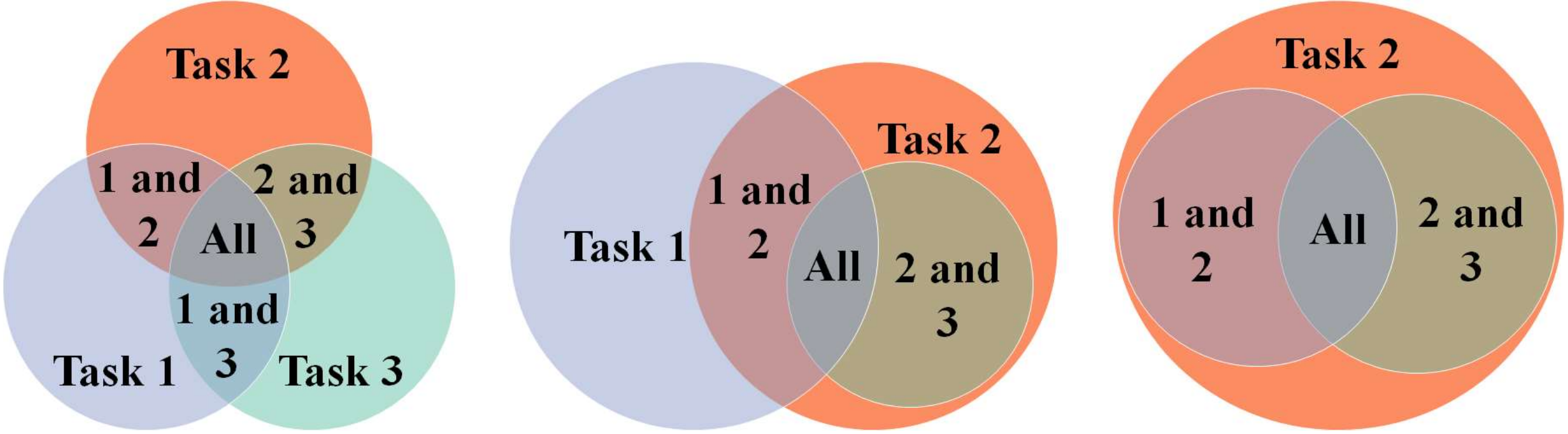}
    \caption{Alternative ways principle tasks can be arranged in grasping models. Note the difference in the overlapping principle tasks. For these grasp models Task 1 refers to Usage, Task 2 to Transfer, and Task 3 is Handover. The different structures represent scenarios where grasp models may lack commonality in poses among tasks. The far left model contains seven zones, $k=7$, while the middle model does not contain ``Task 3 only" nor ``Task 1 and Task 3 only", thus only containing five zones, $k=5$. The right model goes further in also eliminating ``Task 1 only" reducing it to four zones, $k=4$. }
    \label{fig:circ_comp_fig}
\end{figure*}

\subsubsection{Inference Ambiguity}
Due to the planning framework, the resulting poses and solutions are deterministic. The results for the single model comparison assume relatively clean intent inference and will hold three separate cases. The first is a single likely task where the target intent is for Usage(0.9,0.1,0.1). The second is where two likely tasks where the target intents are for Usage and Transfer(0.9,0.9,0.1). Lastly, all three tasks--Usage, Transfer, and Handover--are equally likely(0.9,0.9,0.9). Table \ref{tab:Single_Cup_tab_7x1} shows the comparison of the target probability vector, $v$, and robot probability vector $P_k$. These results show when the scenario with the single likely task is prominent, it primarily will take features from this population, while mostly disregarding other populations. This coincides to how independent model strategies exist. We can also see when more ambiguity is introduced, or the closer the tasks intents are to one another, the better the optimization can handle pulling features from the separate populations to achieve a solution.

\begin{table}[!htbp]
    \centering
    \caption{POSTERIOR PROBABILITY VECTOR COMPARISON FOR THE CUP SINGLE MODEL}
    \begin{tabular}{M{2.8em}|M{3em} M{3em}|M{3em} M{3em}|M{3em} M{3em}}
        %\hline
        H T U&\textbf{Single Task Target}&\textbf{Single Task Final}&\textbf{Two Task Target}&\textbf{Two Task Final}&\textbf{Three Task Target}&\textbf{Three Task Final}\\
        \hline 
        \textbf{0 0 1} & \textbf{0.7933} & \textbf{0.7950} & 0.0817          & 0.0999          & 0.0090          & 0.0108\\
        %\hline 
        \textbf{0 1 0} & 0.0098          & 0.0113          & 0.0817          & 0.0999          & 0.0090          & 0.0105\\
        %\hline 
        \textbf{1 0 0} & 0.0098          & 0.0117          & 0.0010          & 0.0192          & 0.0090          & 0.0049\\
        %\hline 
        \textbf{0 1 1} & 0.0881          & 0.0899          & \textbf{0.7356} & \textbf{0.7538} & 0.0811          & 0.0813\\
        %\hline 
        \textbf{1 0 1} & 0.0881          & 0.0898          & 0.0091          & 0.0000          & 0.0811          & 0.0816\\
        %\hline 
        \textbf{1 1 0} & 0.0011          & 0.0023          & 0.0091          & 0.0272          & 0.0811          & 0.0814\\
        %\hline 
        \textbf{1 1 1} & 0.0098          & 0.0000          & 0.0817          & 0.0000          & \textbf{0.7297} & \textbf{0.7294}\\
        %\hline 
    \end{tabular}
    \label{tab:Single_Cup_tab_7x1}
\end{table}

Table \ref{tab:Single_Cup_tab_7x1} may show the expanded description of the intent, however, we can also reconstruct the initial intent inference, $w$. The reconstruction can show how well the initial intent inference is satisfied for the same intent inference as shown in Table \ref{tab:Single_Cup_tab_3x1}. The comparison of the initial target intent and the final intent given by the final pose is shown in Table \ref{tab:Single_Cup_tab_3x1}. The results of this table reinforce the notion the planner does a better job under greater ambiguity. Along with this analysis it is important to see how well different model structures, shown in Fig. \ref{fig:circ_comp_fig}, impact the final intent inference.  

% Other Table which needs to be reused:
\begin{table}[!htbp]
    \centering
    \caption{PRINCIPLE INTENT INFERENCE COMPARISON FOR THE CUP SINGLE MODEL}
    \begin{tabular}{M{4.06em}|M{2.79em} M{2.79em}|M{2.79em} M{2.79em}|M{2.79em} M{2.79em}}
        \textbf{Initial Intent}&\textbf{Single Task Target}&\textbf{Single Task Final}&\textbf{Two Task Target}&\textbf{Two Task Final}&\textbf{Three Task Target}&\textbf{Three Task Final}\\
        \hline 
        Usage   & 0.9 & 0.9747 & 0.9 & 0.8537 & 0.9 & 0.9031\\
        %\hline 
        Transfer& 0.1 & 0.1035 & 0.9 & 0.8809 & 0.9 & 0.9026\\
        %\hline 
        Handover& 0.1 & 0.1038 & 0.1 & 0.0464 & 0.9 & 0.8973\\
        %\hline 

    \end{tabular}
    \label{tab:Single_Cup_tab_3x1}
\end{table}
%\FloatBarrier

\subsubsection{Model Ambiguity}

Three separate models will be compared as shown in Fig. \ref{fig:circ_comp_fig}. The task listing is as previously discussed, where Task 1 is Usage, Task 2 is Transfer, and Task 3 is Handover. Notice the first model is the standard $k=7$ where each task has unique grasp options, while the second model, $k=5$, does not have any unique classes for ``Handover only" or ``Usage and Handover Only". The last model, $k=4$, takes the second model a step further by also removing the ``Usage only". When all models hold the same case--for instance ``Transfer only"--the results look similar to the single model previously discussed, however, an extreme conditions to examine is where two models ($k=5$ and $k=4$) must accomplish a task which does not exist. In this instance, it is the event for ``Usage and Handover only", where the initial intent vector is (0.9,0.1,0.9). The $k=7$ model is the only one capable of satisfying the true target task. For the $k=5$ and $k=4$ models, Task 3 is a complete subset of Task 2, therefore if Task 3 is inferred then Task 2 must be inferred as well. However, there is conflict with the current initial intent where Task 3 is inferred to be true(0.9), yet Task 2 is inferred to be false(0.1).
In summary, a few key points are needed to clarify the system: 1) even though they have the same initial intent inference input, the descriptor vectors will be different lengths, and thus hold different values; 2) we must note the overlap difference as shown in Fig. \ref{fig:circ_comp_fig}, where the middle model, Task 3 is a complete subset of Task 2. This means any time you complete Task 3, you will automatically complete Task 2. This extension is further shown in the far right model where any time you complete either Task 1 or Task 3 you will complete Task 2 by default.

\begin{table}[!htbp]
    \centering
    \caption{POSTERIOR PROBABILITY VECTOR COMPARISON FOR THE DIFFERENT CUP MODEL}
    \begin{tabular}{M{2.8em}|M{3em} M{3em}|M{3em} M{3em}|M{3em} M{3em}}
        %\hline
        H T U&\textbf{Seven-Zone Target}&\textbf{Seven-Zone Final}&\textbf{Five-Zone Target}&\textbf{Five-Zone Final}&\textbf{Four-Zone Target}&\textbf{Four-Zone Final}\\
        \hline 
        \textbf{0 0 1} & 0.0817          & 0.1049          & 0.4475          & 0.4455          & /               & 0\\
        %\hline 
        \textbf{0 1 0} & 0.0010          & 0.0253          & 0.0055          & 0.0184          & 0.0100          & 0.0129\\
        %\hline 
        \textbf{1 0 0} & 0.0817          & 0.1017          & /               & 0               & /               & 0\\
        %\hline 
        \textbf{0 1 1} & 0.0091          & 0.0069          & 0.0497          & 0.0471          & 0.0900          & 0.0892\\
        %\hline 
        \textbf{1 0 1} & \textbf{0.7356} & \textbf{0.7562} & /               & 0               &/               & 0\\
        %\hline 
        \textbf{1 1 0} & 0.0091          & 0.0049          & 0.0497          & 0.0436          & 0.0900          & 0.0888\\
        %\hline 
        \textbf{1 1 1} & 0.0817          & 0.0000          & 0.4475          & 0.4454          & \textbf{0.8100} & \textbf{0.8091}\\
        %\hline 
    \end{tabular}
    \label{tab:Multi_Cup_tab_7x1}
\end{table}

In Table \ref{tab:Multi_Cup_tab_7x1}, first, the target probability each model should reach is noticeably different and focuses on different regions to achieve the task. Additionally, notice how the different models are able to find a solution which nearly matches the target probability. This demonstrates the reliance the planner has on the model to determine the most appropriate pose. For instance, when looking at the Five-Zone model, the probability vector distribution shows it relies more heavily on the the ``Usage only" (0.4455) and the ``All" (0.4454) zone to pull features. This is due to the initial intent inference containing ambiguity in whether to trust the Transfer or Handover inference, while maintaining the high trust in the Usage inference. Due to the contradicting intent of Transfer and Handover, the intent disambiguation is unsure to believe the more aggressive choice of ``Usage only" task, or to choose the safer option of the ``All" task. Compared to the Four-Zone model which primarily looks at ``the All" zone (0.8091) since any action is considered Task 2. Since the Seven-Zone model already has class for ``Usage and Handover Only" it pulls features from that zone.

\begin{table}[!htbp]
    \centering
    \caption{PRINCIPLE INTENT INFERENCE COMPARISON FOR THE DIFFERENT CUP MODELS}
    \begin{tabular}[width=\columnwidth]{M{4.16em}|M{4.16em}|M{4.16em}|M{4.16em}|M{4.16em}}
        \textbf{Principle Tasks}&\textbf{Initial Intent} &\textbf{Seven-Zone Model}&\textbf{Five-Zone Model}&\textbf{Four-Zone Model}\\
        \hline 
        Usage   & 0.9 & 0.8680 & 0.938 & 0.8983\\
        %\hline 
        Transfer& 0.1 & 0.0371 & 0.5555 & 1.0000\\
        %\hline 
        Handover& 0.9 & 0.8628 & 0.4925 & 0.8979\\
        %\hline 

    \end{tabular}
    \label{tab:Multi_Cup_tab_3x1}
\end{table}

Now that we have established the different zones which influence the overall final pose, the reconstruction of the principle intent can be compared as shown in Table \ref{tab:Multi_Cup_tab_3x1}. Notice how in the Four-Zone model the Transfer task is actually changed to a correct inference of 1.0. This is because the knowledge base of the model understands all grasp poses are a part of the Transfer task. Likewise, the Five-Zone model previously is focusing on two separate zones, the ``Usage only", and ``All". A further inspection shows an inconsistent input inference for the model where Transfer is not supposed to be present while the Handover task says this is more likely. This means the robot knowledge will rely on the more sure Usage task while attempting to find a compromise between using Transfer and Handover features. When comparing the three models in terms of divergence we can use the KL divergence previously discussed. 

\begin{table}[!htbp]
    \centering
    \caption{MODEL COMPARISON FOR BOTH SYMMETRIC AND NONSYMMETRIC DIVERGENCE}

    \begin{tabular}[width=\columnwidth]{M{2.9em}|M{10.19em}|M{10.19em}}
    Model Type& Nonsymmetric & Symmetric \\
    \hline
    Seven-Zone&$\begin{matrix}0&9.21&5.96\\24.04&0&3.06\\33.29&10.02&0\\ \end{matrix}$&$\begin{matrix}0&33.24&39.25\\ 33.24&0&13.08\\39.25&13.08&0\\ \end{matrix}$\\
    \hline
    Five-Zone &$\begin{matrix}0&9.21&12.94\\24.04&0&1.89\\37.17&1.47&0\\ \end{matrix}$&$\begin{matrix}0&33.24&50.11\\33.24&0&3.36\\ 50.11&3.36&0\\ \end{matrix}$\\
    \hline
    Four-Zone &$\begin{matrix}0&4.57&7.40\\31.89&0&1.89\\48.87&1.47&0\\ \end{matrix}$&$\begin{matrix}0&36.46&56.27\\36.46&0&3.36\\56.27&3.36&0\\ \end{matrix}$\\
    \end{tabular}
   \begin{tablenotes}
      \small
      \item The nonsymmetric matrices, the first row uses Usage as the true population, the second row uses Transfer, and the third row is Handover. Likewise, the first column is using Usage as the inference while the second column is Transfer and the third column is Handover. The symmetric matrices follow the similar format; the diagonal terms are added together as stated in Eq. \ref{kl_sym_mat}.
    \end{tablenotes}
    \label{tab:div_table}
\end{table}

The tables show how impacting the model design can change the overall divergence. For instance, when observing the Seven-Zone model to the Five-Zone model (either the nonsymmetric or symmetric matrices), we see the clear decrease in divergence between Transfer and Handover tasks--where the nonsymmetric matrices contain $KL(T||H)_7=3.06$ and $KL(H||T)_7=10.02$ decreases to $KL(T||H)_5=1.89$, and $KL(H||T)_5=1.47$. However, we also eliminated the shared Usage and Handover task, as such, we see an increase in the divergence between these two tasks--where the nonsymmetric matrices of $KL(U||H)_7=5.96$ and $KL(H||U)_7=33.29$ increases to $KL(U||H)_5=12.94$, and $KL(H||U)_5=37.17$. Furthermore, when comparing the Five-Zone and Four- Zone model, we notice no change between Handover and Transfer--where the nonsymmetric matrices contain both $KL(T||H)=1.89$ and $KL(H||T)=1.47$. Interestingly, we can also see a directed decrease in divergence when observing the nonsymmetric matrices. By removing the Usage only component, both the Transfer and Handover tasks do better at inferring the Usage task--where $KL(U||T)_5=9.21$ and $KL(U||H)_5=12.94$ decrease to $KL(U||T)_4=4.57$ and $KL(U||H)_4=7.40$, respectively. However, the Usage data are now worse at predicting the other two tasks--$KL(T||U)_5=24.04$ and $KL(H||U)_5=37.17$ decrease to $KL(T||U)_4=31.89$ and $KL(H||U)_4=48.87$. 
When comparing these models, the Seven-Zone model appear to be the most suitable as it allows for more unique classes and subtle features to pull from when the intent is ambiguous. Additionally, the gain of reducing the divergence levels (increasing model overlap thus increasing model ambiguity) between Transfer and Handover does not seem to out weigh the additional divergence generated with the Usage task. Although, this criteria is highly dependent upon the designers expertise, where some may view the additional divergence generated to be within the acceptable tolerance. Therefore, the recommendation for which model should be used is case dependent (i.e. object dependency, rule/tasks dependency, and robot dependency), however, the approach above is one tool for designers to make the best informed decision in practical scenarios. With this demonstration of the objective results being able to handle the different models, a more extensive look will be observed for different objects.

\subsubsection{Object Ambiguity}

\begin{figure}[!htbp]
   % \centering
    \includegraphics[width=\columnwidth]{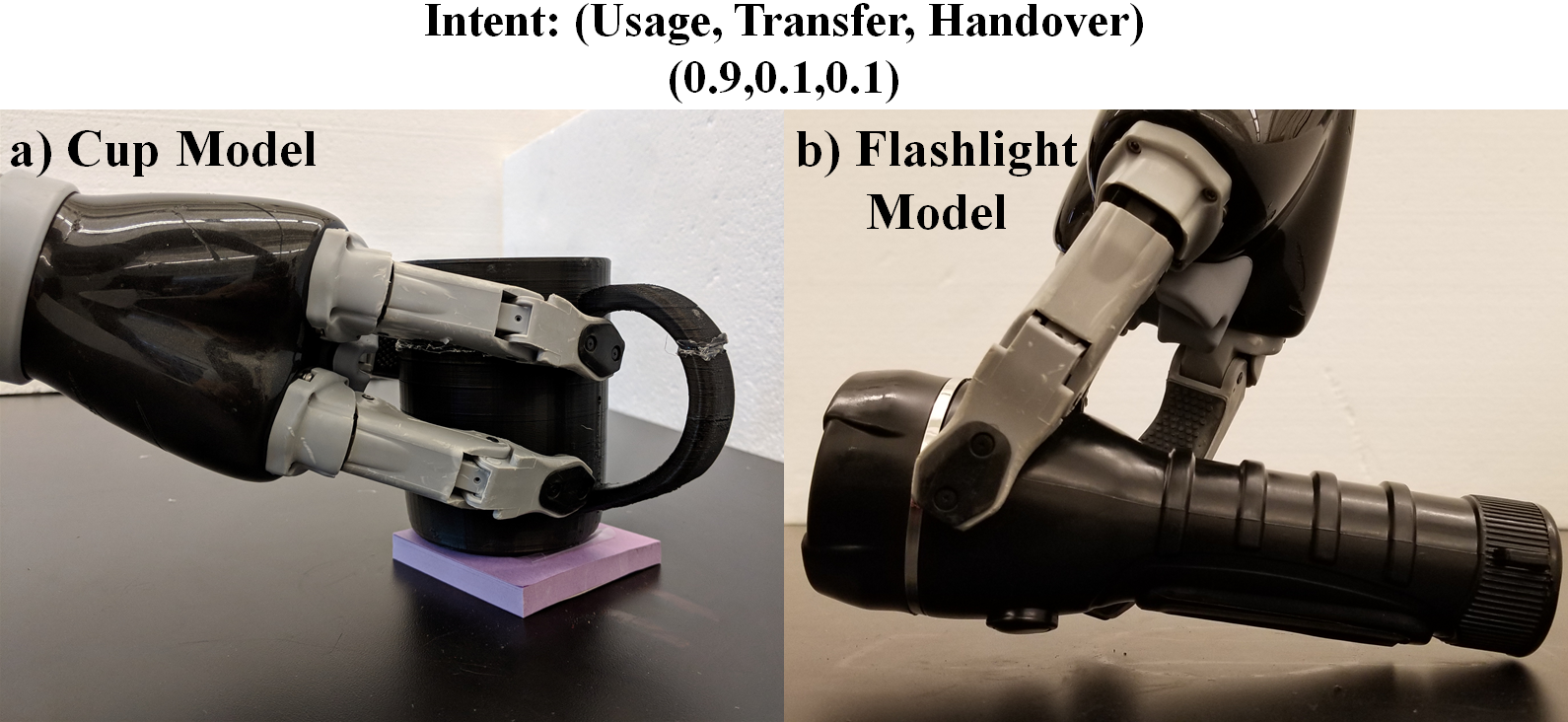}
    \caption{The specific grasp poses generated for different objects from the same initial intent inference. The same intent inference is achieved despite a difference in the model, object, and grasp pose. Further demonstrating the robustness of the planning strategy.}
    \label{fig:mult-_obj_fig}
\end{figure}

For both object comparisons, the Usage task with be used.
Usage grasp poses are fundamentally different for using a cup(drinking) and using a flashlight(shining) as shown in Fig. \ref{fig:mult-_obj_fig}, therefore there are two separate object dependent grasp models created.

The cup grasp dominates the handle area and maintains a palm direction relatively parallel to the table, while the flashlight has a grasp more perpendicular to the table resembling a cup model Transfer only pose (see Table \ref{tab:Cup_subtle_table} for details of subtle motion). 
Therefore, there is not only object differences, but model difference in the preference of the grasp pose. 
Despite these differences, we can still compare these different grasp poses through the intent inference for the same target intent--the task is Usage thus the vector is (0.9,0.1,0.1).

{}
\begin{table}[!htbp]
    \centering
    \caption{PRINCIPLE INTENT INFERENCE COMPARISON FOR DIFFERENT OBJECT MODELS}
    \begin{tabular}[width=\columnwidth]{M{5.2em}|M{5.2em}|M{5.2em}|M{5.2em}}
        \textbf{Principle Tasks}&\textbf{Initial Intent} &\textbf{Cup Model}&\textbf{Flashlight Model}\\
        \hline 
        Usage   & 0.9 & 0.9747 & 0.9760\\
        %\hline 
        Transfer& 0.1 & 0.1035 & 0.1023\\
        %\hline 
        Handover& 0.1 & 0.1038 & 0.1023\\
        %\hline 

    \end{tabular}
    \label{tab:Multi_Object_tab_3x1}
\end{table}
In Table \ref{tab:Multi_Object_tab_3x1}, the reconstructed initial intent is compared to show how well the planner performs despite the different grasp poses. 
The planner is able to take poses fundamentally different from one another and still obtain similar intent inference because of the reliance on the robot grasping model. 
With the inclusive labeling approach, the robot model already learns different poses for different tasks(i.e drink from cup, shine flashlight), thus upon recognizing the object of interest for a specific task it can begin to determine the correct subtle features to extract to produce a matching result. Both the cup and flashlight grasp models could potentially be merged together for a larger model. Where given the target object, the planner can choose the suitable grasp poses for the given intent inference. Furthermore, the analysis to see how well humans can distinguish the intent inference for different objects will also be discussed.

\begin{figure*}[!htbp]
    \centering
    \includegraphics[width=\textwidth]{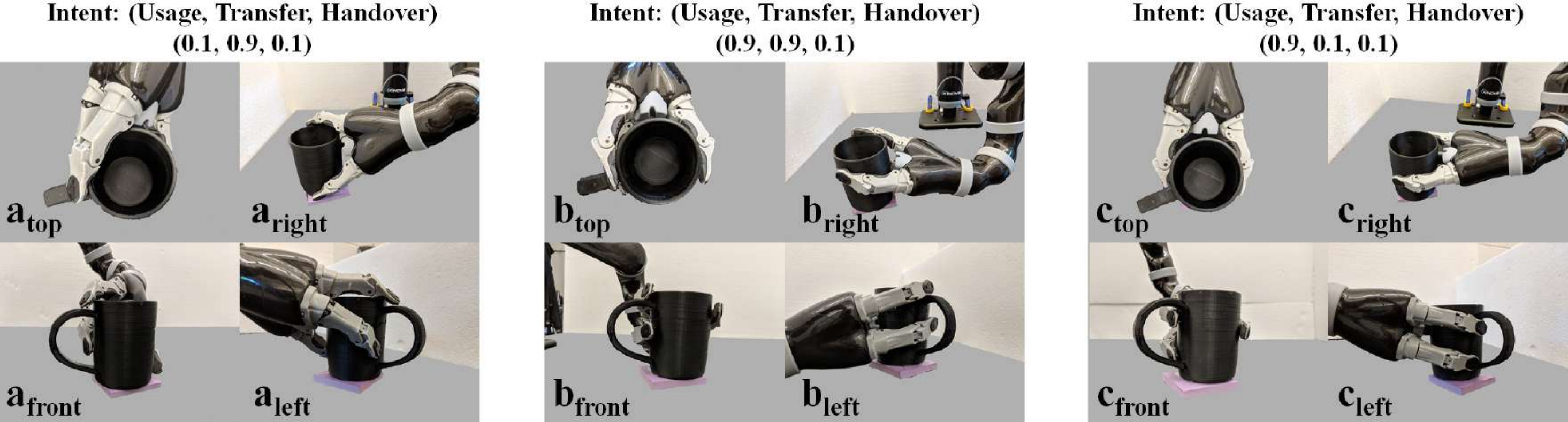}
    \caption{There are three separate grasp poses. a) represents the pose which satisfies the Transfer task and not the other two tasks as seen by the intent provided on the top. This grasp dominates the top of the cup to leave room to place and it is unsuitable for Usage. b) is the pose which our model generated which carries the subtleties to satisfy both Usage and Transfer while not satisfying handover. The robot hand dominates the handle by having a majority of fingers around it, while leaving sufficient room to drink from the top. The palm is higher on the cup. c) is the pose for which satisfies the usage task. The fingers dominate the handle space by having a majority of fingers near the handle, sufficient room on the top for drinking.}
    \label{fig:subjective_cup_model_1}
\end{figure*}
  
\subsection{Grasping Impact}

It is imperative to look at qualitative results to ensure the intent inference is not only an appropriate way to analyze the grasp, but also to ensure the resulting grasps emulate people's expectations. There are two aspects of subjective results which need to be considered, the qualitative results and the human predictability. 

The qualitative results hold two key factors which are to be observed, where the first is by looking at the subtle features of the grasp configuration and the second is to determine if the pose can satisfy the true task. To determine both properties, observe Fig. \ref{fig:subjective_cup_model_1}. Here, two ambiguous intents between Usage and Transfer are used to illustrate the possible shortcomings or possible luck of the independent models. The intent for grasp pose b) is (0.9,0.9,0.1), so there exists difficulty to determine which task to follow. For instance, if the true task is Usage and the independent model approach incorrectly infers the task is Transfer, it will produce the grasp pose a) shown on the left. The subtle differences for this grasp result in a failure for Usage such as the finger covering the top of the cup and leaving insufficient room to drink. If the independent model did correctly infer the Usage task, it would result in grasp c). Where grasp c) appears to be an optimal grasp for drinking, yet it is not optimal for transferring the cup because it lacks palm contact and is a finger dominated grasp which is not a stable configuration. Grasp b) takes features from both Usage and Transfer populations, which generate poses a) and c), to produce a pose which is better equipped to handle both tasks. The palm is close to the cup--resulting in a stable grasp--while also leaving sufficient room to drink. The subtle differences allow a change in the finger placement to leave sufficient room to drink, and palm contact to ensure a stable grasp. These subtle differences of features within a pose allow for a smoother transition of tasks which reduce the risk for inferring the wrong intent. People are keen on picking up subtleties, so it is imperative to account for the ambiguity of intent by creating flexible models which account for it.

Along with the subjective analysis in this manner, a human study--consisting of three of the developers--was conducted. The goal of the study was to analyze how well human predictability aligns with the robot predictability. The developers were to identify 50 poses generated by the model and correctly label the poses for which tasks it could satisfy. This was done twice, once for the Seven-Zone cup model and once for the Seven-Zone flashlight model. Across the three subjects, $73\%$ accuracy was achieved for the cup model, while $76\%$ was achieved for the flashlight model. All subjects appeared to overestimate the amount of tasks a grasp could satisfy. The overestimation could be due to several factors: 1) the subjects were very keen on seeing subtle features such as palm contact to guide their rationale, 2) the poses generated were taking subtle features from relatively unimportant tasks and thus appear to over represent these unimportant tasks, 3) the models created were rather ambiguous, where poses representing distinct classes appear very similar in nature. Future work should be dedicated to a study focusing on this phenomenon with different robots, and different grasp models to see if people consistently overestimate grasp potential. Likewise, further effort should be dedicated to understand the reasoning behind what makes grasp poses legible and transparent. 

\section{Discussion}

Considering ambiguity between different grasp configurations requires the robot models to become continuous rather than discrete models. The continuous model allows more flexibility in grasp configurations to generate predictable and legible poses. Effectively, putting the burden onto the robot to understand and make a pose which is suitable for the human rather than the human learning the associated behavior for specific tasks. Natural expected motion and grasp behavior is imperative for current and future robotic assistant planners where continuous motion is a necessary step to ease mental workload. Considering the overlap of tasks also allow for commonality between the tasks. The commonality can be used to guide the robot agent to a solution when it is unsure of what task is truly correct due to the uncertainty in the human intent inference. Although the final generated pose may be suboptimal, they could potentially be optimal for the human perceived task success. Further, the approach of disambiguation allows the robot to distinguish task success with grasp success. 

It is imperative for planners to start to move from optimal grasp poses for task success to more human centric planners which fulfill the manipulation intent of the operators. The first necessary step is to distinguish the intentions of the human. Disambiguation of the principle tasks occurs through interpreting them as independent, thus transforming the probability to a more descriptive representation. The independent tasks allow for unique combination of the likely tasks. This approach may change the reconstructed intent, yet it is still able to obtain a pose which is suitable for the initial intent inference. This is due to the uncertainty from the human intent inference. There is a need to account for the uncertainty--obtained from human intent and inferencing models--when robotic systems have any level of autonomy during the teleoperation process.

Subtle features play key factors into the success of the autonomy during the teleoperation. The subtle changes are natural to human grasping and are exploited through the common poses to aid in perceived task success. People naturally pick up on the subtleties of hand motion and placement, thus must be considered to generate a natural symbiotic robotic grasp. Likewise, this method can be extended to more objects without significant difference in performance. Objects may hold different important subtleties which are critical for the task success, despite possibly holding the same grasp success configuration. The intent matching is agnostic to object specific grasp models and is created for generalized grasp models. The fundamental grasp configuration for different objects could be different; however, it determines a pose which matches the intent inference the best through manipulating the subtle features unique to the robot model. All together, we account for uncertainty by generating a continuous Bayesian structure from which the final intent distribution closely mimics the human intent.

\section{Conclusion}
The overall approach proves effective due to its practicability compared to the current standards in modeling. The overall approach has demonstrated it is effective in handling intent inference as an input and planning a corresponding grasp. The interpretation of intent inference allows the planner to make a better judgement of which features to utilize, along with  a better explanation of which regions to exploit. The introduction of inclusive labeling to grasp models should allow for more robust, safer, and predictable choices the robot can make in uncertain environments. The strategy has demonstrated it can handle three forms of ambiguity(inference, model, and object ambiguity) which occur in practical scenarios. Lastly, we introduce an approach for designers to quantify their model overlap to evaluate similarity from the inclusive labeling. However, further effort must be put into two areas before the true potential of this work can be utilized. It needs to consider how a person prefers to accomplish the task. This also ties into understanding and generating appropriate ways to analyze manipulation intent. Effort needs to be extended into analyzing the physical discrepancy issue between different hand structures so actions of a robot agent can easily be adapted into a predictable manner to a person. Additionally, learning through demonstration could be applied to the robot model to enhance the understanding of specific human manipulation intent. Despite these necessary next steps, the approach presented is a catalyst to intent-based methods being used in teleoperation. 
\section{Acknowledgments}
This material is based on work supported by the US NSF under grant 1652454. Any opinions, findings, and conclusions or recommendations expressed in this material are those of the authors and do not necessarily reflect those of the National Science Foundation.
\bibliographystyle{IEEEtran}
\bibliography{IEEEabrv,references}

\begin{thebibliography}{10}
\providecommand{\url}[1]{#1}
\csname url@samestyle\endcsname
\providecommand{\newblock}{\relax}
\providecommand{\bibinfo}[2]{#2}
\providecommand{\BIBentrySTDinterwordspacing}{\spaceskip=0pt\relax}
\providecommand{\BIBentryALTinterwordstretchfactor}{4}
\providecommand{\BIBentryALTinterwordspacing}{\spaceskip=\fontdimen2\font plus
\BIBentryALTinterwordstretchfactor\fontdimen3\font minus
  \fontdimen4\font\relax}
\providecommand{\BIBforeignlanguage}[2]{{%
\expandafter\ifx\csname l@#1\endcsname\relax
\typeout{** WARNING: IEEEtran.bst: No hyphenation pattern has been}%
\typeout{** loaded for the language `#1'. Using the pattern for}%
\typeout{** the default language instead.}%
\else
\language=\csname l@#1\endcsname
\fi
#2}}
\providecommand{\BIBdecl}{\relax}
\BIBdecl

\bibitem{Rybarczyk2003ContributionRobot}
Y.~Rybarczyk, E.~Colle, and P.~Hoppenot, ``{Contribution of neuroscience to the
  teleoperation of rehabilitation robot},'' in \emph{IEEE International
  Conference on Systems, Man and Cybernetics}.\hskip 1em plus 0.5em minus
  0.4em\relax Institute of Electrical and Electronics Engineers (IEEE), 8 2003,
  p.~6.

\bibitem{Healey2008SpeculationSurgery}
A.~N. Healey, ``{Speculation on the neuropsychology of teleoperation:
  Implications for presence research and minimally invasive surgery},''
  \emph{Presence: Teleoperators and Virtual Environments}, vol.~17, no.~2, pp.
  199--211, 4 2008.

\bibitem{Li2015ContinuousControl}
Y.~Li, K.~P. Tee, W.~L. Chan, R.~Yan, Y.~Chua, and D.~K. Limbu, ``{Continuous
  Role Adaptation for Human-Robot Shared Control},'' \emph{IEEE Transactions on
  Robotics}, vol.~31, no.~3, pp. 672--681, 6 2015.

\bibitem{Webb2016UsingTeleoperation}
J.~D. Webb, S.~Li, and X.~Zhang, ``{Using visuomotor tendencies to increase
  control performance in teleoperation},'' in \emph{Proceedings of the American
  Control Conference}, vol. 2016-July.\hskip 1em plus 0.5em minus 0.4em\relax
  Institute of Electrical and Electronics Engineers Inc., 7 2016, pp.
  7110--7116.

\bibitem{Song2010LearningModels}
D.~Song, K.~Huebner, V.~Kyrki, and D.~Kragic, ``{Learning task constraints for
  robot grasping using graphical models},'' in \emph{IEEE/RSJ 2010
  International Conference on Intelligent Robots and Systems, IROS 2010 -
  Conference Proceedings}, 2010, pp. 1579--1585.

\bibitem{Song2015Task-BasedInference}
D.~Song, C.~H. Ek, K.~Huebner, and D.~Kragic, ``{Task-Based Robot Grasp
  Planning Using Probabilistic Inference},'' \emph{IEEE Transactions on
  Robotics}, vol.~31, no.~3, pp. 546--561, 6 2015.

\bibitem{Han2016Self-ReflectiveBehaviors}
\BIBentryALTinterwordspacing
F.~Han, C.~Reardon, L.~E. Parker, and H.~Zhang, ``{Self-Reflective Risk-Aware
  Artificial Cognitive Modeling for Robot Response to Human Behaviors},'' in
  \emph{2016 IEEE International Conference on Robotics and Automation
  (ICRA)}.\hskip 1em plus 0.5em minus 0.4em\relax IEEE, 5 2016, pp. 3301--3308.
  [Online]. Available: \url{http://arxiv.org/abs/1605.04934}
\BIBentrySTDinterwordspacing

\bibitem{Lin2000ModelingMotion}
J.~Lin, Y.~Wu, T.~S. Huang, and B.~Institute, ``{Modeling the Constraints of
  Human Hand Motion},'' in \emph{Proceedings Workshop on Human Motion}.\hskip
  1em plus 0.5em minus 0.4em\relax IEEE, 2000, pp. 121--126.

\bibitem{Bowman2019Intent-uncertainty-awareTelemanipulationb}
M.~Bowman, S.~Li, and X.~Zhang, ``{Intent-uncertainty-aware grasp planning for
  robust robot assistance in telemanipulation},'' in \emph{Proceedings - IEEE
  International Conference on Robotics and Automation}, vol. 2019-May, 2019.

\bibitem{Leeper2012StrategiesGrasping}
A.~E. Leeper, K.~Hsiao, M.~Ciocarlie, L.~Takayama, and D.~Gossow, ``{Strategies
  for human-in-the-loop robotic grasping},'' in \emph{ACM/IEEE International
  Conference on Human-Robot Interaction (HRI)}.\hskip 1em plus 0.5em minus
  0.4em\relax ACM/IEEE, 3 2012, p.~1.

\bibitem{Corteville2007Human-inspiredMovements}
B.~Corteville, E.~Aertbelien, H.~Bruyninckx, J.~De~Schutter, and
  H.~Van~Brussel, ``{Human-inspired robot assistant for fast point-to-point
  movements},'' in \emph{Proceedings - IEEE International Conference on
  Robotics and Automation}, 2007, pp. 3639--3644.

\bibitem{Losey2018TrajectoryInteraction}
D.~P. Losey and M.~K. O~Malley, ``{Trajectory Deformations from Physical
  Human-Robot Interaction},'' \emph{IEEE Transactions on Robotics}, vol.~34,
  no.~1, pp. 126--138, 2 2018.

\bibitem{Hirche2012Human-orientedTeleoperation}
S.~Hirche and M.~Buss, ``{Human-oriented control for haptic teleoperation},''
  \emph{Proceedings of the IEEE}, vol. 100, no.~3, pp. 623--647, 3 2012.

\bibitem{Khoramshahi2018AInteraction}
M.~Khoramshahi and A.~Billard, ``{A dynamical system approach to
  task-adaptation in physical human–robot interaction},'' pp. 1--20, 4 2018.

\bibitem{Michelman2002SharedSystem}
P.~Michelman and P.~Allen, ``{Shared autonomy in a robot hand teleoperation
  system},'' in \emph{IEEE/RSJ International Conference on Intelligent Robots
  and Systems}.\hskip 1em plus 0.5em minus 0.4em\relax IEEE, 12 2002, pp.
  253--259.

\bibitem{Kaupp2010Human-robotApproach}
T.~Kaupp, A.~Makarenko, and H.~Durrant-Whyte, ``{Human-robot communication for
  collaborative decision making - A probabilistic approach},'' \emph{Robotics
  and Autonomous Systems}, vol.~58, no.~5, pp. 444--456, 5 2010.

\bibitem{Mulling2015AutonomyManipulation}
\BIBentryALTinterwordspacing
K.~Mulling, A.~Venkatraman, J.-S. Valois, J.~Downey, J.~Weiss, S.~Javdani,
  M.~Hebert, A.~Schwartz, J.~Collinger, and A.~Bagnell, ``{Autonomy Infused
  Teleoperation with Application to BCI Manipulation},'' \emph{Robotics:
  Science and Systems XI}, 7 2015. [Online]. Available:
  \url{http://www.roboticsproceedings.org/rss11/p39.pdf}
\BIBentrySTDinterwordspacing

\bibitem{Abbott2007HapticManipulation}
J.~J. Abbott, P.~Marayong, and A.~M. Okamura, ``{Haptic Virtual Fixtures for
  Robot-Assisted Manipulation},'' in \emph{Robotics Research}.\hskip 1em plus
  0.5em minus 0.4em\relax Springer Berlin Heidelberg, 5 2007, pp. 49--64.

\bibitem{Feygin2002HapticSkill}
D.~Feygin, M.~Keehner, and F.~Tendick, ``{Haptic Guidance: Experimental
  Evaluation of a Haptic Training Method for a Perceptual Motor Skill},'' in
  \emph{Haptic Interfaces for Virtual Environment and Teleoperator Systems.
  HAPTICS}, University of California, Berkeley.\hskip 1em plus 0.5em minus
  0.4em\relax IEEE, 2002, pp. 40--47.

\bibitem{Li2003RecognitionFixtures}
M.~Li and A.~M. Okamura, ``{Recognition of operator motions for real-time
  assistance using virtual fixtures},'' in \emph{Proceedings - 11th Symposium
  on Haptic Interfaces for Virtual Environment and Teleoperator Systems,
  HAPTICS 2003}.\hskip 1em plus 0.5em minus 0.4em\relax Institute of Electrical
  and Electronics Engineers Inc., 2003, pp. 125--131.

\bibitem{Aarno2005AdaptiveTasks}
D.~Aarno, S.~Ekvall, and D.~Kragi{\'{c}}, ``{Adaptive virtual fixtures for
  machine-assisted teleoperation tasks},'' in \emph{Proceedings - IEEE
  International Conference on Robotics and Automation}, vol. 2005, 2005, pp.
  1139--1144.

\bibitem{Dragan2013AControl}
A.~D. Dragan and S.~S. Srinivasa, ``{A policy-blending formalism for shared
  control},'' \emph{International Journal of Robotics Research}, vol.~32,
  no.~7, pp. 790--805, 6 2013.

\bibitem{Huebner2012BADGr-AGRasping}
K.~Huebner, ``{BADGr-A toolbox for box-based approximation, decomposition and
  GRasping},'' \emph{Robotics and Autonomous Systems}, vol.~60, no.~3, pp.
  367--376, 3 2012.

\bibitem{Cutkosky1989OnTasks}
M.~Cutkosky, ``{On grasp choice, grasp models, and the design of hands for
  manufacturing tasks},'' \emph{IEEE Transaction on Robotics and Automation},
  vol.~5, no.~3, pp. 269--279, 1989.

\bibitem{Quispe2016GraspingPlanning}
\BIBentryALTinterwordspacing
A.~H. Quispe, H.~B. Amor, H.~Christensen, and M.~Stilman, ``{Grasping for a
  Purpose: Using Task Goals for Efficient Manipulation Planning},'' \emph{arXiv
  eprint}, 3 2016. [Online]. Available: \url{http://arxiv.org/abs/1603.04338}
\BIBentrySTDinterwordspacing

\bibitem{Song2011MultivariateGrasping}
\BIBentryALTinterwordspacing
D.~Song, C.~H. Ek, K.~Huebner, and D.~Kragic, ``{Multivariate Discretization
  for Bayesian Network Structure Learning in Robot Grasping},'' in \emph{IEEE
  International Conference on Robotics and Automation}.\hskip 1em plus 0.5em
  minus 0.4em\relax IEEE, 2011, pp. 1944--1950. [Online]. Available:
  \url{http://www.csc.kth.se/cvap,}
\BIBentrySTDinterwordspacing

\bibitem{Song2013PredictingInteractions}
D.~Song, N.~Kyriazis, I.~Oikonomidis, C.~Papazov, A.~Argyros, D.~Burschka, and
  D.~Kragic, ``{Predicting human intention in visual observations of
  hand/object interactions},'' in \emph{Proceedings - IEEE International
  Conference on Robotics and Automation}, 2013, pp. 1608--1615.

\bibitem{Zhou2007SolvingClustering}
Z.~H. Zhou and M.~L. Zhang, ``{Solving multi-instance problems with classifier
  ensemble based on constructive clustering},'' \emph{Knowledge and Information
  Systems}, vol.~11, no.~2, pp. 155--170, 2 2007.

\bibitem{Cheng2013ComparingClassifiers}
\BIBentryALTinterwordspacing
J.~Cheng and R.~Greiner, ``{Comparing Bayesian Network Classifiers},''
  \emph{CoRR}, 1 2013. [Online]. Available:
  \url{http://arxiv.org/abs/1301.6684}
\BIBentrySTDinterwordspacing

\bibitem{Tzima2015InducingSystems}
\BIBentryALTinterwordspacing
F.~A. Tzima, M.~Allamanis, A.~Filotheou, and P.~A. Mitkas, ``{Inducing
  Generalized Multi-Label Rules with Learning Classifier Systems},''
  \emph{CoRR}, 12 2015. [Online]. Available:
  \url{http://arxiv.org/abs/1512.07982}
\BIBentrySTDinterwordspacing

\bibitem{Zhang2010Multi-labelDependency}
\BIBentryALTinterwordspacing
M.-L. Zhang and K.~Zhang, ``{Multi-label learning by exploiting label
  dependency},'' in \emph{Proceedings of the 16th ACM SIGKDD international
  conference on Knowledge discovery and data mining - KDD '10}.\hskip 1em plus
  0.5em minus 0.4em\relax New York, New York, USA: ACM Press, 2010, p. 999.
  [Online]. Available:
  \url{http://dl.acm.org/citation.cfm?doid=1835804.1835930}
\BIBentrySTDinterwordspacing

\bibitem{John2013EstimatingClassifiers}
\BIBentryALTinterwordspacing
G.~H. John and P.~Langley, ``{Estimating Continuous Distributions in Bayesian
  Classifiers},'' \emph{CoRR}, 2 2013. [Online]. Available:
  \url{http://arxiv.org/abs/1302.4964}
\BIBentrySTDinterwordspacing

\bibitem{Heckerman2008ANetworks}
D.~Heckerman, ``{A Tutorial on Learning With Bayesian Networks},'' in
  \emph{Innovations in Bayesian Networks: Theory and Applications}, D.~E.
  Holmes and L.~C. Jain, Eds.\hskip 1em plus 0.5em minus 0.4em\relax Springer
  Berlin Heidelberg, 2008, pp. 33--82.

\bibitem{Waltz2006AnSteps}
R.~A. Waltz, J.~L. Morales, J.~Nocedal, and D.~Orban, ``{An Interior Algorithm
  for Nonlinear Optimization That Combines Line Search and Trust Region
  Steps},'' \emph{Mathematical Programming}, vol. 107, no.~3, pp. 391--408,
  2006.

\bibitem{BrandtPetersenMichaelSyskindPedersen2012TheCookbook}
K.~Brandt Petersen Michael Syskind~Pedersen, B.~Baxter, B.~Templeton,
  C.~Rish{\o}j, C.~Schr{\"{o}}ppel, D.~Boley, D.~L. Theobald,
  E.~Hoegh-Rasmussen, E.~Karseras, G.~Martius, G.~Casteel, J.~Larsen,
  J.~Bin~Gao, J.~Struckmeier, K.~Dedecius, K.~T. Abou-Moustafa, K.~Strimmer,
  L.~Christiansen, L.~Kai~Hansen, L.~Wilkinson, L.~He, L.~Thibaut, M.~Froeb,
  M.~Hubatka, M.~Bar{\~{a}}o, O.~Winther, P.~Sakov, S.~Hattinger, T.~Pedersen,
  V.~Sima, and V.~Rabaud, ``{The Matrix Cookbook},'' 2012.

\bibitem{Hershey2007AppoximatingModels}
J.~R. Hershey and P.~A. Olsen, ``{Appoximating The Kullback Leibler Divergence
  Between Gaussian Mixture Models},'' in \emph{IEEE International Conference on
  Acoustics, Speech and Signal Processing - ICASSP}.\hskip 1em plus 0.5em minus
  0.4em\relax IEEE, 2007, pp. IV--317--IV--320.

\bibitem{Nocedal2006NumericalOptimization}
J.~Nocedal and S.~Wright, \emph{{Numerical optimization}}.\hskip 1em plus 0.5em
  minus 0.4em\relax Springer Science {\&} Business Media, 2006.

\bibitem{Boggs2000SequentialOptimization}
P.~T. Boggs and J.~W. Tolle, ``{Sequential quadratic programming for
  large-scale nonlinear optimization},'' \emph{Journal of Computational and
  Applied Mathematics}, vol. 124, no. 1-2, pp. 123--137, 2000.

\end{thebibliography}
\end{document}